\documentclass[10pt,twocolumn,letterpaper]{article}

\usepackage{cvpr}
\usepackage{times}
\usepackage{epsfig}
\usepackage{graphicx}
\usepackage{amsmath}
\usepackage{amssymb}

\usepackage[pagebackref=true,breaklinks=true,letterpaper=true,colorlinks,bookmarks=false]{hyperref}

 \cvprfinalcopy 

\def\cvprPaperID{726} 

\ifcvprfinal\fi
\begin{document}

\title{Agile Amulet: Real-Time Salient Object Detection with Contextual Attention}

\author{Pingping Zhang$^{\dagger}$\quad Luyao Wang$^{\dagger}$\quad Dong Wang$^{\dagger}$\quad Huchuan Lu$^{\dagger}$\quad Chunhua Shen$^{\ddagger}$\thanks{Prof. Shen is the corresponding author.}\\
$^{\dagger}$Dalian University of Technology, P. R. China\quad\quad $^{\ddagger}$University of Adelaide, Australia\\
{\tt\small \{jssxzhpp,luyaow\}@mail.dlut.edu.cn\quad \{wdice,lhchuan\}@dlut.edu.cn\quad chunhua.shen@adelaide.edu.au}
}
\maketitle

\begin{abstract}
This paper proposes an Agile Aggregating Multi-Level feaTure framework (\textbf{Agile Amulet}) for salient object detection.
The Agile Amulet builds on previous works to predict saliency maps using multi-level convolutional features.
Compared to previous works, Agile Amulet employs some key innovations to improve training and testing speed while also increase prediction accuracy.
More specifically, we first introduce a contextual attention module that can rapidly highlight most salient objects or regions with contextual pyramids.
Thus, it effectively guides the low-layer convolutional feature learning and tells the backbone network where to look.
The contextual attention module is a fully convolutional mechanism that simultaneously learns complementary features and predicts saliency scores at each pixel.
In addition, we propose a novel method to aggregate multi-level deep convolutional features.
As a result, we are able to use the integrated side-output features of pre-trained convolutional networks alone, which significantly reduces the model parameters leading to a model size of \textbf{67} MB, about half of Amulet.
%
%
Compared to other deep learning based saliency methods, Agile Amulet is of much lighter-weight, runs faster (\textbf{30} fps in real-time) and achieves higher performance on seven public benchmarks in terms of both quantitative and qualitative evaluation.
%
\end{abstract}

\section{Introduction}
Salient object detection, which aims to identify the most conspicuous objects or regions in an image, is one of the fundamental problems in computer vision community.
It can serve as the first step of many object-related applications, such as pattern classification~\cite{sharma2012discriminative,zhang2015saliency}, instance retrieval~\cite{wang2002semantics,he2012mobile,gao20123,cheng2014salientshape}, sematic segmentation~\cite{achanta2008salient,donoser2009saliency}, image thumbnailing~\cite{marchesotti2009framework}, visual tracking~\cite{mahadevan2009saliency,hong2015online,cane2016saliency} and person re-identification~\cite{zhao2013unsupervised,zhao2013person}.
In general, salient object detection methods should be fast, accurate, and able to recognize and localize a wide variety of objects.
Since the introduction of deep convolutional neural networks (CNNs), salient object detection frameworks
have become more and more accurate~\cite{zhao2015saliency,wang2015deep,li2015visual,lee2016deep,liu2016dhsnet,Li2016Deep,DSSalCVPR2017}.
However, most of saliency methods are still constrained to low speed and high complexity, which drags them on wide-ranging applications.
\begin{figure}
\centering
\begin{tabular}{@{}c}
\includegraphics[width=1\linewidth,height=5.6cm]{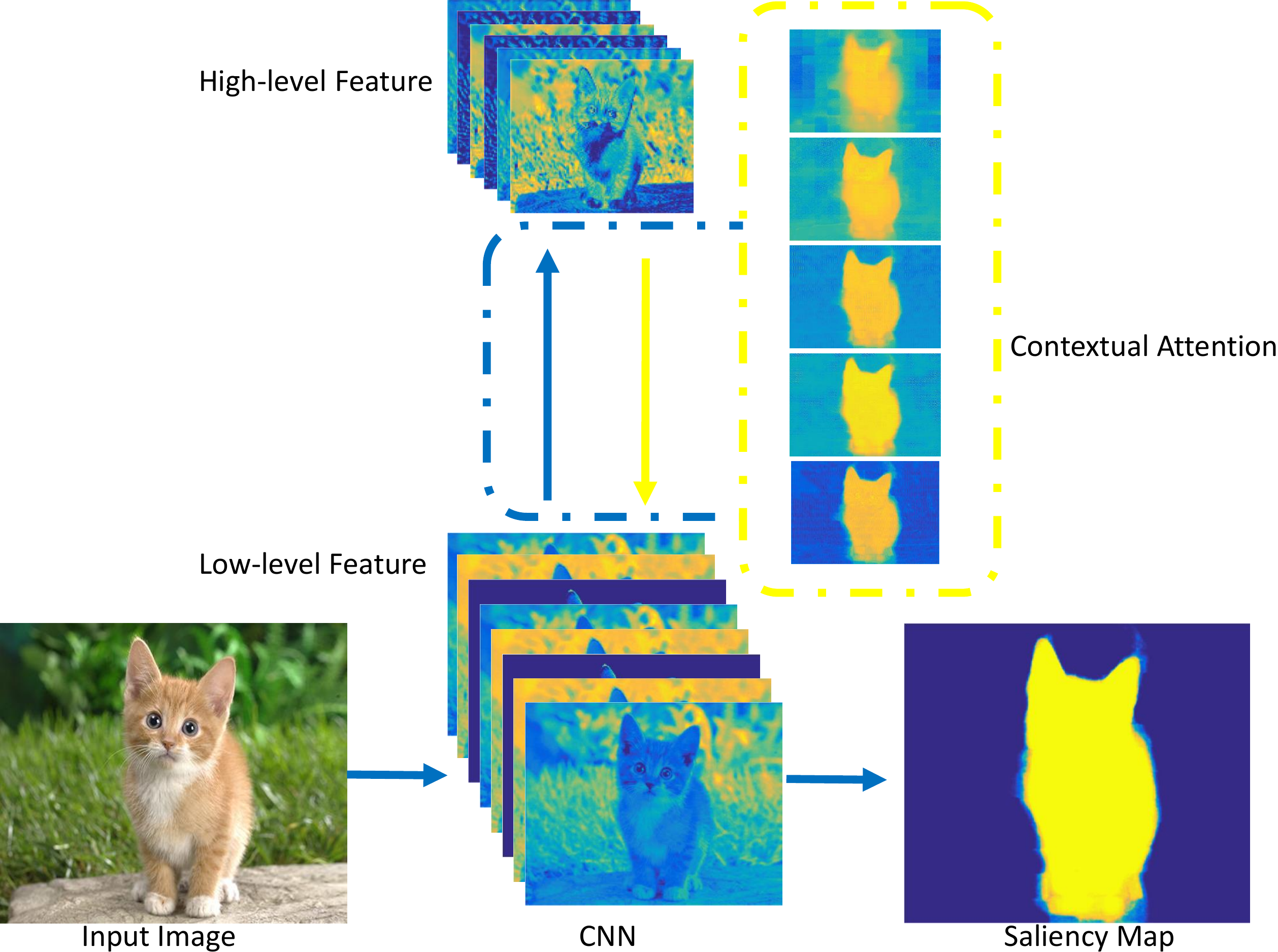}\ \\
\end{tabular}
\vspace{2mm}
\caption{Contextual attention module. It can be inserted between any CNN layers.
An attention pyramid with varied contexts is generated based on multi-level feature maps and previous predictions, which highlights most salient objects or regions of the input image.
During the training procedure, the contextual attention also guides the low-layer feature learning and forces the backbone network to focus on the informative object regions.
\label{fig:sca}}
\vspace{-4mm}
\end{figure}

In this paper, we simplify the over-designed frameworks and the overload training progress of state-of-the-art deep CNN-based salient object detectors~\cite{liu2016dhsnet,DSSalCVPR2017,zhang2017amulet}.
To this end, we introduce a \emph{contextual attention module} that can rapidly highlight salient objects or regions with stacked contextual pyramids.
Thus it is able to guide low-layer convolutional feature learning and tell the backbone network where to look closely, as shown in Fig.~\ref{fig:sca}.
We also propose a new aggregating multi-level feature method that significantly reduces model parameters and complexity. 
The resulting approach, named \textbf{Agile Amulet}, can train a very deep detection network (\eg, VGG-16~\cite{simonyan2014very}) five times faster than Amulet~\cite{zhang2017amulet}, run twice faster (30 \emph{fps} in real-time) at test-time, and achieve a higher state-of-the-art performance on seven public saliency detection benchmarks.
\subsection{Drawbacks of DHS, DSS and Amulet}
The deeply supervised learning-based saliency methods, \eg, DHS~\cite{liu2016dhsnet}, DSS~\cite{DSSalCVPR2017} and Amulet~\cite{zhang2017amulet}, have achieved excellent salient object detection accuracy by using a pre-trained deep CNN and multiple supervised losses.
However, these methods still have several notable drawbacks:
\begin{itemize}
\vspace{-0.5mm}
\item
CNN architectures are over-designed.
More specifically, DHS utilizes several recurrent convolutional layer (RCL)~\cite{liang2015recurrent} to capture image local context information, however, each RCL inherently incorporates multiple recurrent connections into each convolutional layer.
In DSS method, a series of short connections are densely introduced for combining features in deeper and shallower layers.
Intuitively, Amulet integrates multi-level convolutional features using resolution-based feature combination structures, which need all convolutional features as inputs.
While effective, the above designs are very redundancy and computationally inefficient.
\vspace{-0.5mm}
\item
Network training is expensive in space and time.
For deep feature extraction, all aforementioned methods use the pre-trained VGG-16 network~\cite{simonyan2014very}.
This process takes about 0.05 GPU-seconds and 5 gigabytes of storage for each image with 256$\times$256 resolution.
Because more computationally inefficient convolutional layers are introduced into their frameworks, training requires larger storage space and more running time.
\vspace{-0.5mm}
\item
Saliency prediction is slower than real-time (25 \emph{fps}). At test-time, with a high-end GTX Titan X GPU and 256$\times$256 resolution images, DHS runs at about 22.5 \emph{fps}. Amulet runs 16.2 \emph{fps}. DSS runs 2.5 \emph{fps}.
When only using CPUs, these methods performs very slowly (about 10 s/image).
Unsurprisingly, this speed limitation hampers them on real-world applications, especially on the embedded devices.
\end{itemize}
\subsection{Main Contributions}
In this work, we propose a novel salient object detection algorithm that overcomes the disadvantages of existing methods, \eg, DHS~\cite{liu2016dhsnet}, DSS~\cite{DSSalCVPR2017} and Amulet~\cite{zhang2017amulet}, while improves performance in both speed and accuracy.

We term this method \textbf{Agile Amulet} because it is similar with Amulet~\cite{zhang2017amulet} but comparatively lighter-weight, more flexible and faster to train and test.
The key difference is that we introduce a contextual attention module into the feature learning stage.
It is able to highlight salient objects or regions, thus guide the backbone network to focus on the object-related features.
Besides, we propose a novel aggregating feature method that is quite different from the one used in Amulet~\cite{zhang2017amulet}.
Our method uses an iterative process to aggregate the multi-level features, which significantly reduces the parameters and computations.
To overcome the prediction inconsistency of the deeply supervised learning, we propose a recursive prediction method, that can progressively improve results upon previous predictions.
Compared to other deep learning based methods, our method has several advantages as follows:

1. Higher saliency detection performance is achieved on seven large-scale salient object detection datasets.

2. Contextual attention is used for fast salient region extraction and effective low-layer feature learning guidance.

3. Model size and complexity are significantly reduced, using our new aggregating multi-level feature method.

4. Testing can be real-time and use multi-context predictions without result fusing.
\section{Related Work}
\subsection{Salient Object Detection}
In recent years, deep learning based methods, especially CNNs, have delivered
remarkable performance in salient object detection.
For example, Wang \etal~\cite{wang2015deep} use two deep neural networks to integrate local pixel estimation and global proposal search for salient object detection.
Li \etal~\cite{li2015visual} predict the saliency degree of each superpixel by using multi-scale features in multiple generic CNNs.
Zhao \etal~\cite{zhao2015saliency} take global and local context into account, and predict saliency in a multi-context deep CNN framework.
%
Lee \etal~\cite{lee2016deep} propose to encode low-level distance map and high-level sematic features of deep CNNs for salient object detection.
Liu \etal~\cite{liu2016dhsnet} propose a deep hierarchical saliency network to progressively refine saliency maps.
In addition, using multiple deep CNNs, Li \etal~\cite{Li2016Deep} design a pixel-level stream and a segment-level stream to produce more accurate saliency predictions.
Wang \etal~\cite{wang2016saliency} propose deep recurrent fully convolutional networks (FCNs) to incorporate saliency priors and stage-wisely refine the prediction.
Hou \etal~\cite{DSSalCVPR2017} propose a new saliency method by introducing short connections to the HED architecture~\cite{xie2015holistically}.
Zhang \etal~\cite{zhang2017learning} employ a convolutional encoder-decoder network with R-dropout modules to acquire accurate saliency maps.
Zhang \etal~\cite{zhang2017amulet} propose a bidirectional learning method to adaptively aggregate multi-level convolutional features for salient object detection.
In addition, Wang \etal~\cite{Wang_2017_ICCV} develop a multi-stage refinement mechanism and augment plain deep neural networks with a global context module for saliency detection.
\begin{figure*}
\begin{center}
\includegraphics[width=0.95\linewidth,height=8.0cm]{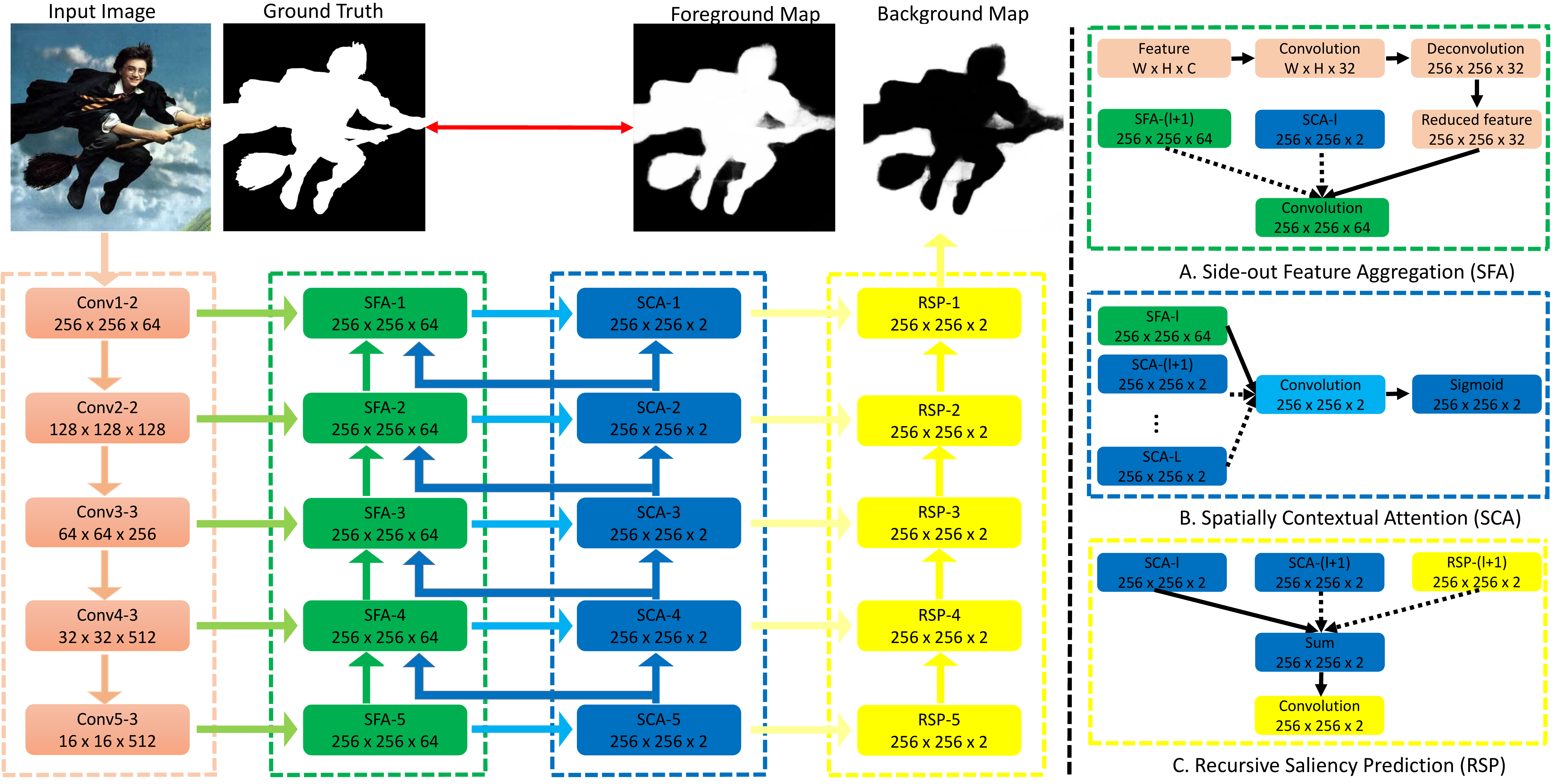}
\end{center}
\vspace{-4mm}
\caption{An overview of our approach. Left: The framework of our proposed model based on the VGG-16 model~\cite{simonyan2014very}.
Right: The details of Side-output Feature Aggregation (SFA), Spatially Contextual Attention (SCA) and Recursive Saliency Prediction (RSP) modules are illustrated in (A), (B) and (C), respectively.
Each box is considered as a component.
The solid arrows show the feed-forward information stream, while the dotted arrows mean that specific operations maybe not appear in corresponding components.}
\label{fig:framework}
\vspace{-5mm}
\end{figure*}
\subsection{Spatial Context and Visual Attention}
Spatial context is known to be very useful for improving performance on detection and segmentation tasks~\cite{mottaghi2014role}.
As for dense labeling tasks, it is often ambiguous in the presence of only local information.
However, these tasks become much simpler if contextual information, from large receptive fields, is available.
For instance, Liu \etal~\cite{liu2015parsenet} add global context to plain FCNs~\cite{long2015fully}
for semantic segmentation, using global average pooling.
The approach is simple, but significantly increases the performance of baseline networks.
Yu \etal~\cite{yu2015multi} use dilated convolutions to aggregate multi-scale contextual information.
%
%
They show that the presented context module increases the accuracy of semantic segmentation systems.
To get richer context information, Zhao \etal~\cite{zhao2016pyramid} propose a pyramid pooling module (PPM), which exploit the capability of different-region context information.
The context representation is effective to produce high quality results on the scene parsing task.
Chen \etal~\cite{chen2017rethinking} also design context-based spatial pyramid pooling modules which employ dilated convolution to capture multi-scale context by adopting multiple dilated rates.
These works illustrate that reasonable context information can help dense labeling tasks, \eg, salient object detection.

Another useful method is visual attention mechanism.
Visual attention models have been successfully adopted in many computer vision tasks, including
object recognition~\cite{mnih2014recurrent,ba2014multiple}, fine-grained image classification~\cite{sermanet2014attention,lin2015bilinear}, image caption~\cite{jin2015aligning,anderson2017bottom} and visual question answering (VQA)~\cite{chen2015abc,xu2016ask,lu2016hierarchical,yang2016stacked,anderson2017bottom}.
In most of works, visual attention is modeled as a region sequence in an image.
In general, an recurrent neural network (RNN) model is utilized to predict the next attention region based on the location and visual features of current attention regions.
In contrary to them, we build visual attention on the backbone network's outputs with variable context and low-level complementary features, which are potential saliency cues and helpful for salient object detection.
Our proposed method can guide low-layer convolutional feature learning and tell the backbone network where to look, \ie, focuses on the most salient objects.
With our new aggregating feature method, the contextual attention can significantly reduce model parameters and improve training and testing speed.
%
%
\section{Agile Amulet Approach}
The overall framework of Agile Amulet is illustrated in Fig.~\ref{fig:framework}.
Our Agile Amulet consists of four components: (1) the multi-level feature extraction part (red box), (2) the side-output feature aggregation part (green box), (3) the spatially contextual attention part (blue box) and (4) the recursive saliency prediction part (yellow box).
%
%
The rest of this section will describe each component of our Agile Amulet framework in detail.
\subsection{Multi-level Feature Extraction}
For multi-level feature extraction, we uniformly resize a raw image $I$ into 256$\times$256$\times$3 pixels, then we utilize a deep CNN pre-trained on the 1000-class ImageNet classification challenge dataset~\cite{imagenet_cvpr09}, \ie , VGG-16~\cite{simonyan2014very} or ResNet-50~\cite{he2016deep} to extract the multi-level feature maps $\textbf{f}_{I}$:
\begin{equation}
  \label{equ:equ1}
\begin{aligned}
  \textbf{f}_{I} = (\textbf{f}^{1}(I),...,\textbf{f}^{l}(I),...., \textbf{f}^{L}(I)) = CNN(I),
\end{aligned}
\end{equation}
where $l\in [1, L]$ is the level of deep features. $\textbf{f}^{l}(I)$ is represented as a $N^{l}\times N^{l} \times D^{l}$ feature tensor.
For notional simplicity, we subsequently drop the dependence $I$ and only consider the feature representations.
The \emph{CNN} is the VGG-16 or ResNet-50 model.
For the VGG-16 model, we follow Amulet~\cite{zhang2017amulet} and take the features $\textbf{f}_{I}$ from the front-end convolutional layers (\ie, \emph{conv1-2, conv2-2, conv3-3, conv4-3 and conv5-3}), which retain spatial information of input images.
For the ResNet-50 model, we choose the features $\textbf{f}_{I}$ from the \emph{conv1, res2c, res3d, res4f and res5c} layers.
It is worthy to note that:
(1) The side-output features ($\textbf{f}^{1}, \textbf{f}^{2},..., \textbf{f}^{L}$) essentially are different visual descriptions of input images, which may have different resolutions $N^{l}$ and channels $D^{l}$.
(2) Each pixel in the feature tensor $\textbf{f}^{l}$ corresponds to a large region (receptive field) of the input images.
With the enlarged receptive fields of deep convolutional layers, contextual information is implicitly exploited as regional features.
(3) Similar to Amulet, it is also possible to use other layers or deep CNNs as backbone networks, \eg , VGG-19~\cite{simonyan2014very}, ResNet-101/152~\cite{he2016deep} or DenseNet~\cite{huang2017densely} for the multi-level feature extraction.
\subsection{Side-output Feature Aggregation}
Leveraging the hierarchy features of a deep CNN can boost the detection performance.
However, as mentioned in Subsection 1.1, most of existing works introduce complex convolutional modules to combine features in deep layers and shallow layers.
Those modules inevitably need more computation and run-time.
In contrast to them, here we only use the side-output features of backbone networks and integrate them in a more efficient way.
To this end, we propose a simple yet extremely efficient aggregating method to transform each level features ($\textbf{f}^{l}$) to a dimension-reduced tensor that has the same spatial resolution as the input images.
In particular, we use an iterative process to aggregate the side-output features for a more compact representation.
Formally, the resulting output $\textbf{g}^{l}$ at level $l$ becomes
\begin{equation}
  \label{equ:equ2}
\begin{aligned}
  \textbf{g}^{l} = \phi^{l}([\textbf{w}^{l}_{u}*_{s}\textbf{w}^{l}_{r}*\textbf{f}^{l},\textbf{g}^{l+1},\textbf{a}^{l+1}]),
\end{aligned}
\end{equation}
where [...] represents the concatenation operation.
$\phi$ is defined as the batch-normalization (BN)~\cite{badrinarayanan2015segnet2}, followed by the ReLU activation.
$*$ and $*_{s}$ are the regular convolution operator and de-convolution operator with a stride $s$, respectively.
With parameters $\textbf{w}^{l}_{r}$ and $\textbf{w}^{l}_{u}$, the side-output feature $\textbf{f}^{l}$ can first be reduced to a low-dimension tensor, then up-sampled to the spatial size of the input image.
$\textbf{a}^{l+1}$ is the contextual attention map at level $l+1$. We will elaborate it in the following subsection.
The proposed iterative aggregation pattern in Equ.~\ref{equ:equ2} strongly encourages the reuse of high-level aggregated features and incorporates new complementary low-layer features in the architecture.
Compared to existing aggregation methods~\cite{liu2016dhsnet,DSSalCVPR2017,zhang2017amulet}, our method has less parameters and the output dimension of each layer can be significantly reduced.
In addition, with the guidance of our contextual attention, the aggregation can focus on the features of salient objects or regions instead of the overall feature maps.
Thus, the complexity and size of our model is rather small while more superior performance is achieved.
\subsection{Spatially Contextual Attention}
In general salient objects do not appear in isolation.
They are always surrounded by a related background (\eg, sky and playground) and likely to coexist with other objects.
These contexts provide valuable information to discriminate them from the background.
In addition, salient objects usually occupy a large part of the image and draw human attention.
In light of these facts, we incorporate contextual attention into our feature aggregation and network learning to force the backbone network focus on most salient objects or regions and reduce the negative influence of background.

Considering the higher layer captures more larger context regions and encodes more specific object information, we generate the attention mask from high-level features to low-level features.
As shown in Fig.~\ref{fig:framework} (B), we add a prediction-aware convolutional layer after the low-layer aggregated features.
These additional layers use aggregated features with varied context to generate an attention map,
\begin{equation}
  \label{equ:equ3}
\textbf{a}^{l}=
\left\{
\begin{aligned}
&\sigma(\textbf{w}^l_a*[\textbf{g}^l,\textbf{a}^{l+1}]+\textbf{b}^l_a),l<L\\
&\sigma(\textbf{w}^l_a*\textbf{g}^l+\textbf{b}^l_a), l = L
\end{aligned}
\right.
\end{equation}
where $\sigma(x)=\frac{1}{1+e^{-x}}$ is the sigmoid function. $\textbf{w}^l_a$ and $\textbf{b}^l_a$ are the learnable attention weight and bias, respectively.
Unlike existing methods~\cite{mnih2014recurrent,sermanet2014attention,lin2015bilinear,yang2016stacked,anderson2017bottom}, our proposed contextual attention maps are derived from coarse saliency predictions, \ie, $\textbf{a}^{l}$ approaches to the ground truth of the input image.
When making a new prediction, it is easy to interpret and provide insights into where the network should look at closely.
The attention is supervised by both bottom-up and top-down cues.
The values of the contextual attention map $\textbf{a}^l$ are between 0 and 1, representing the importance of the corresponding regions in the original image and feature maps.
In addition, as shown in Fig.~\ref{fig:framework} (B) and Equ.~\ref{equ:equ3}, the contextual attention map is concatenated with the aggregated low-level feature maps instead of multiplying the input image or low-level features for the following reasons:

1) Multiplying the attention maps with the input image causes fake edges, which may lead to wrong saliency predictions.
Instead, we find that concatenating with the aggregated low-level feature maps can avoid this problem in both training and testing phase.

2) Multiplying the attention maps with low-level features weakens or discards most of useful features, leading to the over-fitting problem on small datasets.
With concatenation, the features are still kept and can be recaptured for the detailed and robust prediction.

\textbf{From single mask to attention pyramids.}
The above method is effective, however, the specific-level attention mask only relies on fixed context information, which limits its ability of detecting multiple objects with varied scales.
To remedy this problem, we leverage the pyramidal shape of ConvNets' context and build a contextual attention pyramid that has rich context at all scales.
Formally, we stack the generated contextual attention maps from the top level to the current level by
\begin{equation}
  \label{equ:equ4}
\textbf{a}^{l}=
\left\{
\begin{aligned}
&\sigma(\textbf{w}^l_a*[\textbf{g}^l,\textbf{a}^{l+1},...,\textbf{a}^{L}]+\textbf{b}^l_a),l<L\\
&\sigma(\textbf{w}^l_a*\textbf{g}^l+\textbf{b}^l_a), l = L
\end{aligned}
\right.
\end{equation}
The resulting attention maps are based on specific-level convolutional features and all available context that has rich semantic information and attains robustness for complex scenes.
We will demonstrate the effectiveness of this new attention pyramid in the experimental section.
\subsection{Recursive Saliency Prediction}
As reflected in~\cite{liu2016dhsnet,DSSalCVPR2017,zhang2017amulet}, it appears inconsistency if we use multiple predictions, where certain predictions on their own can sometimes provide superior results than the final fused output.
Instead, we propose a more straightforward recursive prediction method.
As illustrated in Fig.~\ref{fig:framework} (C), we combine predictions from the current and higher levels using simple addition prior and attention masks.
Formally, our model predicts the saliency map by
\begin{equation}
  \label{equ:equ5}
\textbf{s}^{l}=
\left\{
\begin{aligned}
&\textbf{w}^l_s*(\textbf{a}^{l}+\textbf{a}^{l+1}+\textbf{s}^{l+1})+\textbf{b}^l_s,l<L\\
&\textbf{w}^l_s*\textbf{a}^l+\textbf{b}^l_s, l = L.
\end{aligned}
\right.
\end{equation}
As initial predictions can be negative or positive values, the Equ.~\ref{equ:equ5} actually force the saliency classifier weights $\textbf{w}^l_s$
to improve results upon previous predictions, by recursively adding to or subtracting appropriate information from the corresponding predictions.
This recursive prediction allows our model to avoid the inconsistent outputs.
\section{Training and Testing}
Though described separately in Section 3, our whole framework is trained with image pairs in an end-to-end way.
During the testing, given an image, our method can directly produce its final prediction using the aggregated feature maps and the contextual attention.

\textbf{Network Training.}
Suppose there are $N$ training samples $S=\{(X_n,Y_n)\}^{N}_{n=1}$, and $X_n =\{x^n_j,j = 1,...,T\}$ and $Y_n = \{y^n_j,j = 1,...,T\}$ are the input image and the binary ground-truth with $T$ pixels, respectively.
$y^n_j = 1$ denotes the foreground pixel and $y^n_j = 0$ denotes the background pixel.
For notional simplicity, we subsequently drop the subscript $n$ and consider each image independently.
In addition, we denote $\textbf{W}$ as the parameters of the backbone network. $\theta^l = (\textbf{w}^l_{fa}, \textbf{w}^l_{ca}, \textbf{w}^l_{rp})$ is the parameter of the feature aggregation part, the contextual attention part, and the recursive prediction part at level $l$, respectively.
%
Following~\cite{DSSalCVPR2017,zhang2017amulet}, we employ the cross-entropy loss as the objective function, however, we do not have the fused term because our network design progressively improve upon the previous predictions.
Our objective function is expressed as
\begin{align}
  \mathcal{L}(\textbf{W},\theta)=\sum_{l=1}^{L}\alpha_l\mathcal{L}_l(\textbf{W},\theta^l),
  \label{equ:equ6}
\end{align}
\begin{equation}
  \label{equ:equ7}
\begin{aligned}
  \mathcal{L}_l(\textbf{W},\theta^l)= - \beta \sum_{j\in Y_{+}} \text{log~Pr}(y_{j}=1|X;\textbf{W},\theta^l)\\
-(1-\beta)\sum_{j\in Y_{-}} \text{log~Pr}(y_{j}=0|X;\textbf{W},\theta^l),
\end{aligned}
\end{equation}
where $\alpha_l$ is the loss weight to balance each loss term. For simplicity and fair comparison, we set $\alpha_l=1$ as in~\cite{xie2015holistically,DSSalCVPR2017,zhang2017amulet}.
The class-balancing weight $\beta = |Y_{-}|/|Y|$, $1-\beta = |Y_{+}|/|Y|$, and $|Y_{+}|$ and $|Y_{-}|$ denote the foreground and background pixel number, respectively.
Following~\cite{zhang2017amulet}, we use the softmax classifier to evaluate the prediction scores:
\begin{align}
\text{Pr}(y_{j}=1|X;\textbf{W},\theta^l) = \frac{e^{\textbf{s}^l_1}}{e^{\textbf{s}^l_0}+e^{\textbf{s}^l_1}},
  \label{equ:equ8}
\end{align}
\vspace{-8mm}
\begin{align}
\text{Pr}(y_{j}=0|X;\textbf{W},\theta^l) = \frac{e^{\textbf{s}^l_0}}{e^{\textbf{s}^l_0}+e^{\textbf{s}^l_1}},
  \label{equ:equ9}
\end{align}
where $\textbf{s}^l_0$ and $\textbf{s}^l_1$ are the predicted values of each pixel of the input image.
The above loss function (Equ.~\ref{equ:equ6}) is continuously differentiable, so the standard stochastic gradient descent (SGD) method~\cite{krizhevsky2012imagenet} can be used to obtain the optimal parameters.
See Section 5.1 for detailed hyper-parameters and experimental settings.

\textbf{Forward Testing.}
As described in Subsection 3.4, our network progressively improves the saliency prediction upon previous high-level ones.
Therefore, we can simply use the lowest level prediction as our final saliency map.
Specifically, given an image, we only need to compute the foreground confidence at $l=1$, \ie, $\textbf{S}=\sigma(\textbf{s}^1)$.
This simple saliency inference is realized with minimal complexity, requiring fewer computations than other methods.
\section{Experiments}
In this section, we extensively evaluate our proposed method on seven public datasets and report the runtime.
The experimental results demonstrate that our method is very superior on saliency detection in both accuracy and speed.
\subsection{Experimental Setup}
\textbf{Datasets.}
To train our network, we adopt the MSRA10K dataset~\cite{ChengPAMI}, which contains 10,000 images with pixel-wise saliency annotations.
Most of the images in this dataset have one salient object, the diversity of images is limited.
Thus, we augment this dataset by random cropping, mirror reflection and rotation techniques ($0^{\circ}, 90^{\circ}, 180^{\circ}, 270^{\circ}$), producing 120,000 training images totally.

For the performance evaluation, we adopt seven public saliency detection datasets as follows:

\textbf{DUT-OMRON}~\cite{yang2013saliency}. This dataset has 5,168 high quality images. Each image in this dataset has one or more salient objects with relatively complex background.

\textbf{DUTS-TE}~\cite{wang2017learning}. This dataset is the test set of currently largest saliency detection benchmark (DUTS)~\cite{wang2017learning}. It contains 5,019 images with high quality pixel-wise annotations.

\textbf{ECSSD}~\cite{yan2013hierarchical}. This dataset contains 1,000 natural images, in which many semantically meaningful and complex structures are included.

\textbf{HKU-IS}~\cite{li2015visual}. This dataset has 4,447 images with high quality pixel-wise annotations.
Images of this dataset are well chosen to include multiple disconnected salient objects or objects touching the image boundary.

\textbf{PASCAL-S}~\cite{li2014secrets}. This dataset is generated from the classical PASCAL VOC dataset~\cite{Everingham2010ThePV} and contains 850 natural images with segmentation-based masks.

\textbf{SED}~\cite{borj2015salient}. This dataset has two independent subsets, \ie, \textbf{SED1} and \textbf{SED2}.
\textbf{SED1} has 100 images each containing only one salient object, while \textbf{SED2} has 100 images each containing two salient objects.

\textbf{SOD}~\cite{jiang2013salient}. This dataset has 300 images, in which many images contain multiple objects either with low contrast or touching the image boundary.

\textbf{Implementation Details.}
We implement our approach based on the Caffe toolbox~\cite{jia2014caffe}.
We train and test our approach in a quad-core PC machine with an i5-6600 CPU and an NVIDIA Titan 1080 GPU (with 8G memory).
We train models using augmented images from the MSRA10K dataset.
We do not use validation set and train the model until its training loss converges.
The input image is resized such that it has $256\times256\times3$ pixels.
The parameters of multi-level feature extraction layers are initialized from the VGG-16 model~\cite{simonyan2014very} or ResNet-50~\cite{he2016deep}.
For other layers, we initialize the weights by the ``Xavier'' method~\cite{glorot2010understanding}.
During the training, we use standard SGD method~\cite{krizhevsky2012imagenet} with batch size 8, momentum 0.9 and weight decay 0.0005.
We set the base learning rate to 1e-8 and decrease the learning rate by 10\% when training loss reaches a flat.
The training process converges after 200k iterations.
%
%

\textbf{Evaluation Metrics.}
We use four metrics to evaluate the performance of different saliency detection algorithms, including the widely used precision-recall (PR) curves, F-measure, mean absolute error (MAE)~\cite{borji2015salient} and recently proposed S-measure~\cite{fan2017structure}.
%
%
The PR curve of a dataset demonstrates the mean precision and recall of saliency maps at different thresholds.
The F-measure is a weighted mean of average precision and average recall, calculated by
\vspace{-.5mm}
\begin{align}
  F_{\eta} =\frac{(1+\eta^2)\times Precision\times Recall}{\eta^2\times Precision \times Recall}.
    \label{equ:equ11}
\end{align}
\vspace{-0.5mm}
We set $\eta^2$ to be 0.3 to weigh precision more than recall as suggested in~\cite{yan2013hierarchical}~\cite{wang2015deep}~\cite{borji2015salient}~\cite{yang2013saliency}.
%

The above overlapping-based evaluations usually give higher score to methods which assign high saliency score to salient pixel correctly.
%
%
For fair comparisons, we also calculate the mean absolute error (MAE) by
\vspace{-0.5mm}
\begin{align}
MAE = \frac{1}{W\times H}\sum_{x=1}^{W}\sum_{y=1}^{H}|S(x,y)-G(x,y)|,
  \label{equ:equ12}
\end{align}
\vspace{-0.5mm}
where $W$ and $H$ are the width and height of the input image. $S(x,y)$ and $G(x,y)$ are the pixel values of the saliency map and the binary ground truth at $(x,y)$, respectively.

To evaluate the spatial structure similarities of saliency maps, we also calculate the S-measure (More details appear in~\cite{fan2017structure}.), defined as
\vspace{-0.5mm}
\begin{align}
S_{\lambda} = \lambda*S_{o}+(1-\lambda)*S_{r},
  \label{equ:equ13}
\end{align}
\vspace{-0.5mm}
where $\lambda \in [0,1]$ is the balance parameter. $S_{o}$ and $S_{r}$ are the object-aware and region-aware structural similarity, respectively. We set $\lambda=0.5$ as suggested by the authors.
\subsection{Experimental Results}
%
%
%
\subsubsection{Saliency Detection Results}
%
We compare our proposed algorithm with other 14 state-of-the-art ones, including 10 deep learning based algorithms (Amulet~\cite{zhang2017amulet}, DCL~\cite{Li2016Deep}, DHS~\cite{liu2016dhsnet}, DS~\cite{Li2016DeepSaliency}, DSS~\cite{DSSalCVPR2017}, ELD~\cite{lee2016deep}, LEGS~\cite{wang2015deep}, MDF~\cite{li2015visual}, RFCN~\cite{wang2016saliency}, UCF~\cite{zhang2017learning}) and 4 conventional algorithms (BL~\cite{tong2015bootstrap}, BSCA~\cite{qin2015saliency}, DRFI~\cite{jiang2013salient}, DSR~\cite{li2013saliency}).
For fair comparison, we use either
the implementations with recommended parameter settings
or the saliency maps provided by the authors.
\begin{figure*}
\begin{center}
\begin{tabular}{@{}c@{}c@{}c@{}c}
\includegraphics[width=0.24\linewidth,height=3.85cm]{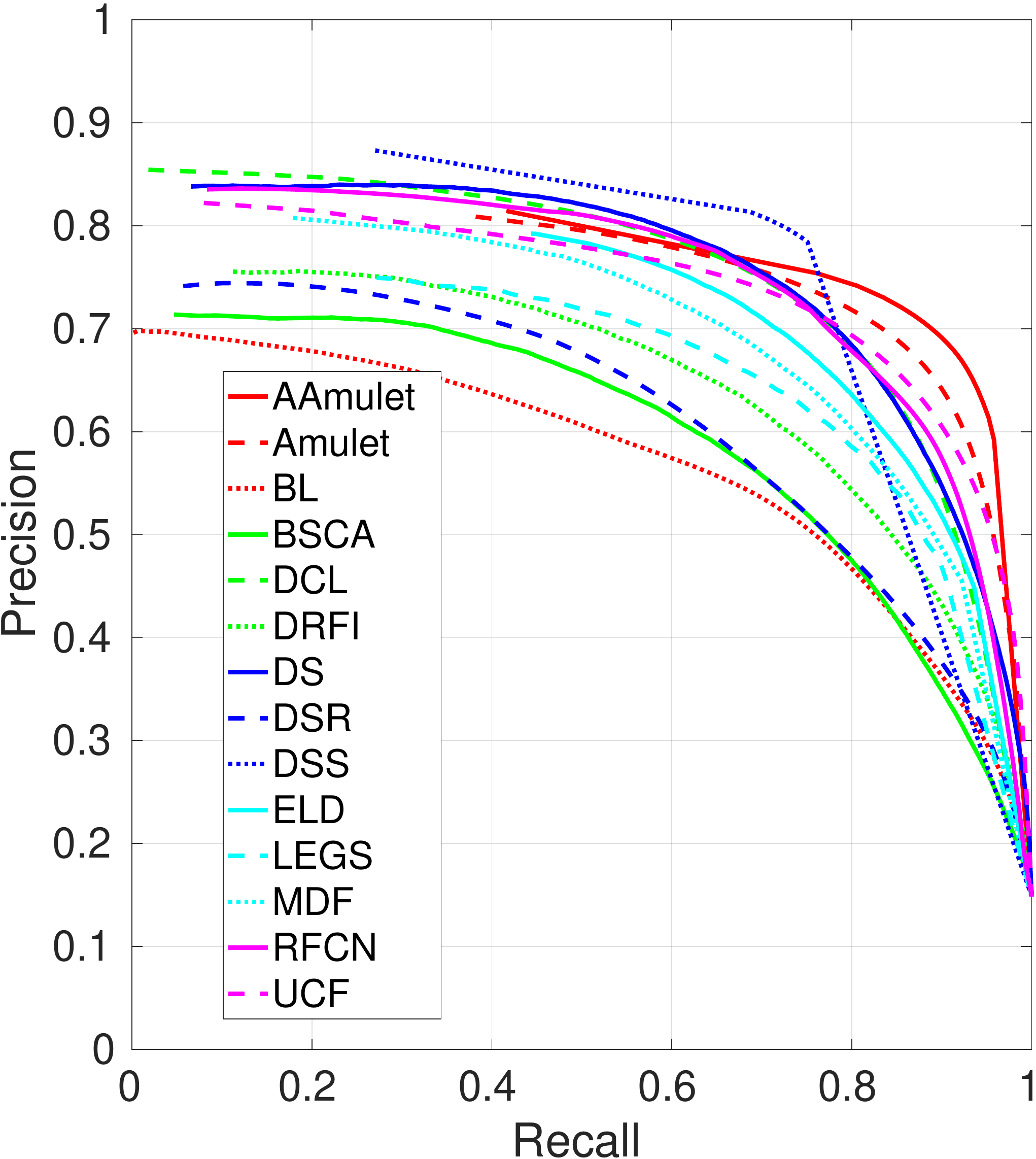} \ &
\includegraphics[width=0.24\linewidth,height=3.85cm]{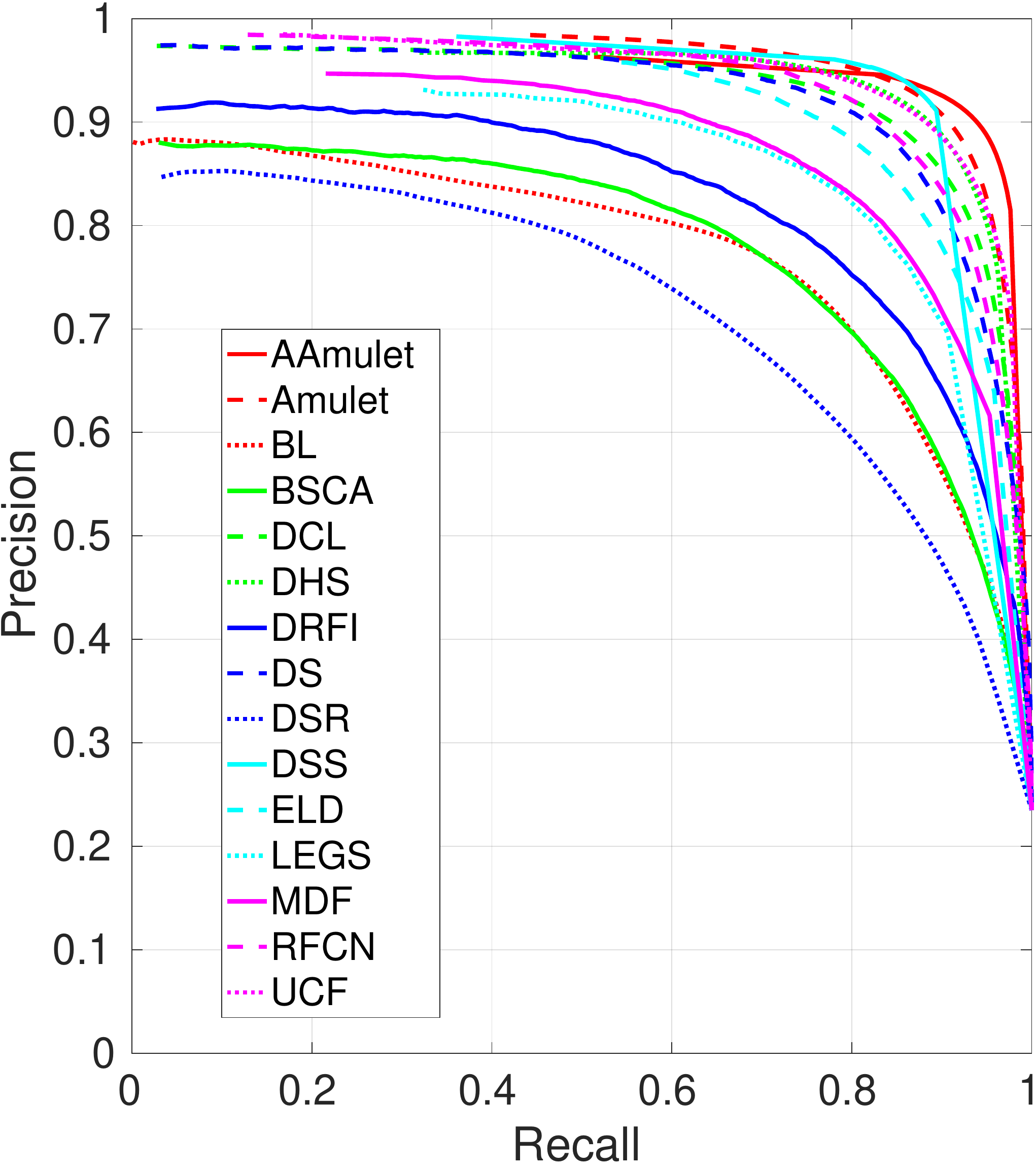} \ &
\includegraphics[width=0.24\linewidth,height=3.85cm]{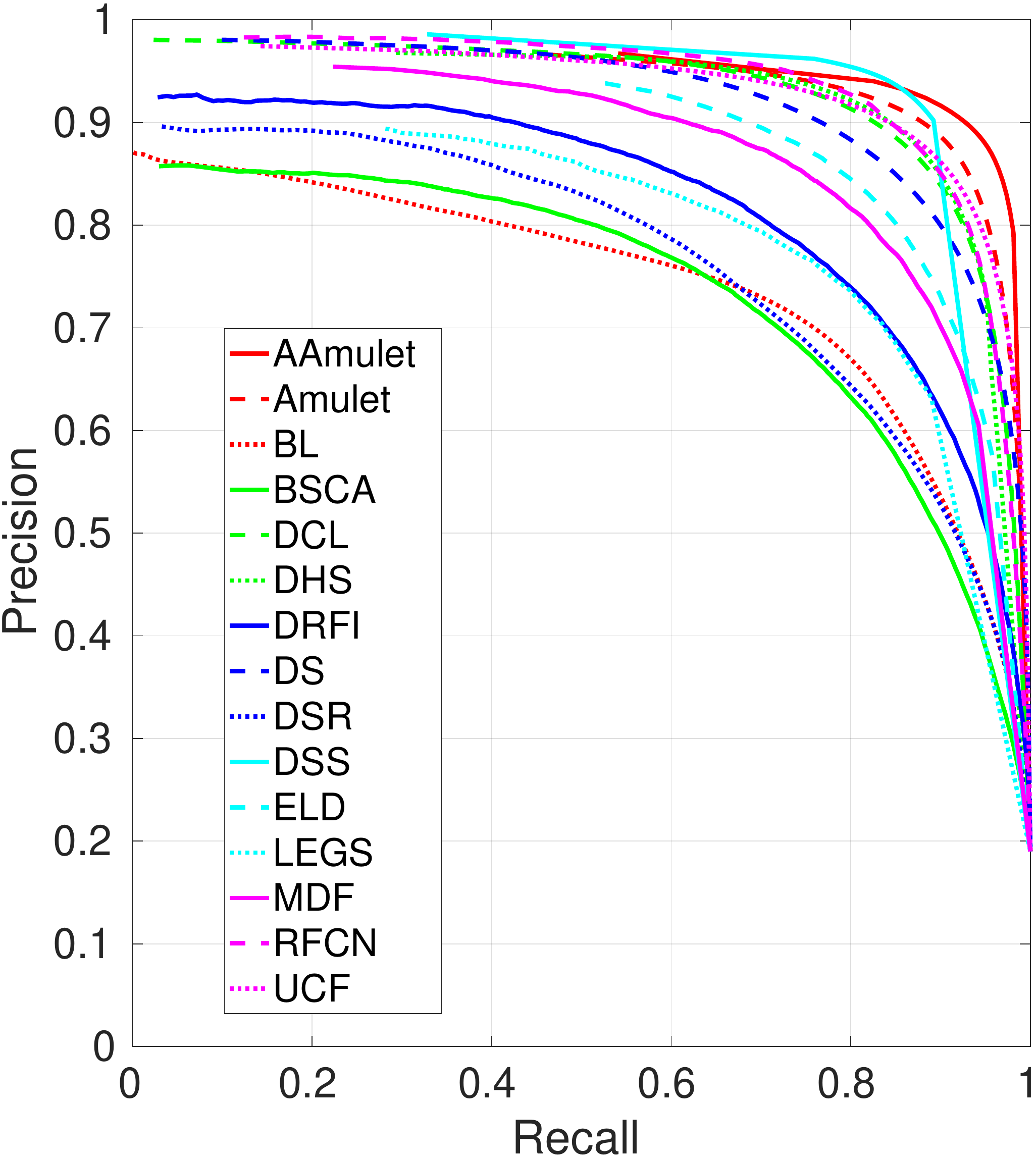} \ &
\includegraphics[width=0.24\linewidth,height=3.85cm]{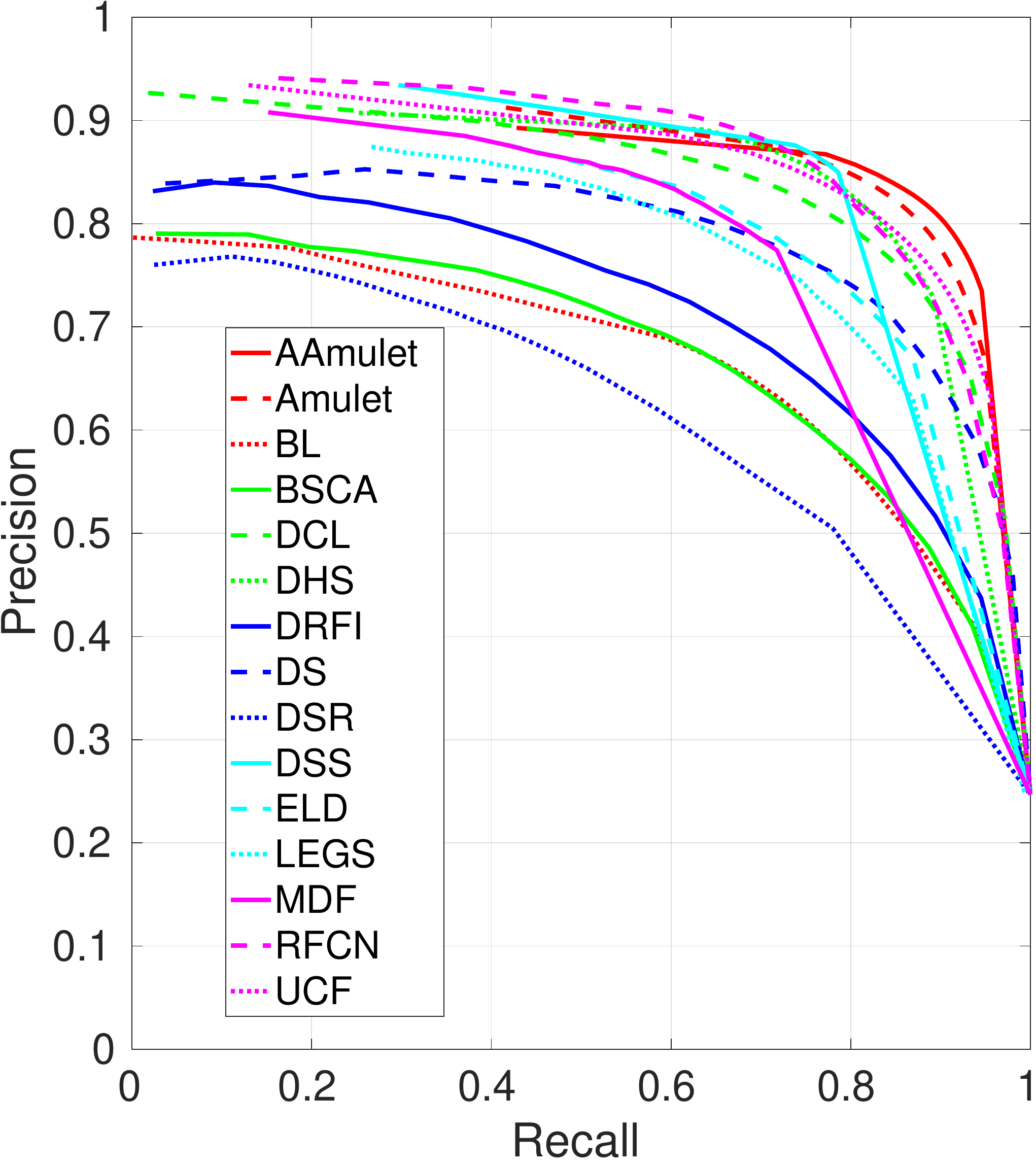} \ \\
 {\small(a) DUT-OMRON} & {\small(b) ECSSD} & {\small(c) HKU-IS} & {\small(d) PASCAL-S}\\
\\
\end{tabular}
\vspace{-4mm}
\caption{The PR curves of the proposed algorithm and other state-of-the-art methods.
\label{fig:PR-curve}}
\end{center}
\end{figure*}
\setlength{\tabcolsep}{5pt}
\begin{table*}
\vspace{-5mm}
\begin{center}
\doublerulesep=0.6pt
\begin{tabular}{|c|c|c|c|c|c|c|c|c|c|c|c|c|c|c|c|c|c|c|c|c|c|c|c|c|||c|c|c|c|c|c|c|c|||}
\hline
\multicolumn{4}{|c|}{}
&\multicolumn{6}{|c|}{DUT-OMRON~\cite{yang2013saliency}}
&\multicolumn{6}{|c|}{ECSSD~\cite{yan2013hierarchical}}
&\multicolumn{6}{|c|}{HKU-IS~\cite{li2015visual}}
&\multicolumn{6}{|c|}{PASCAL-S~\cite{li2014secrets}}
\\
\hline
\hline
\multicolumn{4}{|c|}{Methods}
&\multicolumn{2}{|c|}{$F_\eta$}&\multicolumn{2}{|c|}{$MAE$}&\multicolumn{2}{|c|}{$S_\lambda$}
&\multicolumn{2}{|c|}{$F_\eta$}&\multicolumn{2}{|c|}{$MAE$}&\multicolumn{2}{|c|}{$S_\lambda$}
&\multicolumn{2}{|c|}{$F_\eta$}&\multicolumn{2}{|c|}{$MAE$}&\multicolumn{2}{|c|}{$S_\lambda$}
&\multicolumn{2}{|c|}{$F_\eta$}&\multicolumn{2}{|c|}{$MAE$}&\multicolumn{2}{|c|}{$S_\lambda$}
\\
\hline
\hline
\multicolumn{4}{|c|}{\textbf{AAmulet}}
&\multicolumn{2}{|c|}{\textcolor[rgb]{0,1,0}{0.691}}&\multicolumn{2}{|c|}{\textcolor[rgb]{0,1,0}{0.076}}&\multicolumn{2}{|c|}{\textcolor[rgb]{1,0,0}{0.782}}
&\multicolumn{2}{|c|}{\textcolor[rgb]{0,1,0}{0.887}}&\multicolumn{2}{|c|}{\textcolor[rgb]{1,0,0}{0.049}}&\multicolumn{2}{|c|}{\textcolor[rgb]{1,0,0}{0.902}}
&\multicolumn{2}{|c|}{\textcolor[rgb]{0,1,0}{0.861}}&\multicolumn{2}{|c|}{\textcolor[rgb]{1,0,0}{0.038}}&\multicolumn{2}{|c|}{\textcolor[rgb]{1,0,0}{0.891}}
&\multicolumn{2}{|c|}{\textcolor[rgb]{0,1,0}{0.794}}&\multicolumn{2}{|c|}{\textcolor[rgb]{1,0,0}{0.092}}&\multicolumn{2}{|c|}{\textcolor[rgb]{1,0,0}{0.832}}
\\
\multicolumn{4}{|c|}{\textbf{Amulet}~\cite{zhang2017amulet}}
&\multicolumn{2}{|c|}{0.647}&\multicolumn{2}{|c|}{0.098}&\multicolumn{2}{|c|}{\textcolor[rgb]{0,1,0}{0.771}}
&\multicolumn{2}{|c|}{\textcolor[rgb]{0,0,1}{0.868}}&\multicolumn{2}{|c|}{\textcolor[rgb]{0,0,1}{0.059}}&\multicolumn{2}{|c|}{\textcolor[rgb]{0,1,0}{0.894}}
&\multicolumn{2}{|c|}{0.843}&\multicolumn{2}{|c|}{\textcolor[rgb]{0,0,1}{0.050}}&\multicolumn{2}{|c|}{\textcolor[rgb]{0,1,0}{0.886}}
&\multicolumn{2}{|c|}{\textcolor[rgb]{0,0,0}{0.768}}&\multicolumn{2}{|c|}{0.098}&\multicolumn{2}{|c|}{\textcolor[rgb]{0,1,0}{0.820}}
\\
\multicolumn{4}{|c|}{\textbf{DCL}~\cite{Li2016Deep}}
&\multicolumn{2}{|c|}{\textcolor[rgb]{0,0,1}{0.684}}&\multicolumn{2}{|c|}{0.157}&\multicolumn{2}{|c|}{0.743}
&\multicolumn{2}{|c|}{0.829}&\multicolumn{2}{|c|}{0.149}&\multicolumn{2}{|c|}{0.863}
&\multicolumn{2}{|c|}{0.853}&\multicolumn{2}{|c|}{0.136}&\multicolumn{2}{|c|}{0.859}
&\multicolumn{2}{|c|}{0.714}&\multicolumn{2}{|c|}{0.181}&\multicolumn{2}{|c|}{0.791}
\\
\multicolumn{4}{|c|}{\textbf{DHS}~\cite{liu2016dhsnet}}
&\multicolumn{2}{|c|}{--}&\multicolumn{2}{|c|}{--}&\multicolumn{2}{|c|}{--}
&\multicolumn{2}{|c|}{0.867}&\multicolumn{2}{|c|}{0.601}&\multicolumn{2}{|c|}{0.884}
&\multicolumn{2}{|c|}{\textcolor[rgb]{0,0,1}{0.854}}&\multicolumn{2}{|c|}{0.053}&\multicolumn{2}{|c|}{0.869}
&\multicolumn{2}{|c|}{\textcolor[rgb]{0,0,1}{0.778}}&\multicolumn{2}{|c|}{\textcolor[rgb]{0,1,0}{0.095}}&\multicolumn{2}{|c|}{\textcolor[rgb]{0,0,1}{0.807}}
\\
\multicolumn{4}{|c|}{\textbf{DS}~\cite{Li2016DeepSaliency}}
&\multicolumn{2}{|c|}{0.603}&\multicolumn{2}{|c|}{0.120}&\multicolumn{2}{|c|}{0.741}
&\multicolumn{2}{|c|}{0.826}&\multicolumn{2}{|c|}{0.122}&\multicolumn{2}{|c|}{0.821}
&\multicolumn{2}{|c|}{0.787}&\multicolumn{2}{|c|}{0.077}&\multicolumn{2}{|c|}{0.854}
&\multicolumn{2}{|c|}{0.659}&\multicolumn{2}{|c|}{0.176}&\multicolumn{2}{|c|}{0.739}
\\
\multicolumn{4}{|c|}{\textbf{DSS}~\cite{DSSalCVPR2017}}
&\multicolumn{2}{|c|}{\textcolor[rgb]{1,0,0}{0.740}}&\multicolumn{2}{|c|}{\textcolor[rgb]{1,0,0}{0.063}}&\multicolumn{2}{|c|}{\textcolor[rgb]{0,0,1}{0.764}}
&\multicolumn{2}{|c|}{\textcolor[rgb]{1,0,0}{0.904}}&\multicolumn{2}{|c|}{\textcolor[rgb]{0,1,0}{0.052}}&\multicolumn{2}{|c|}{0.882}
&\multicolumn{2}{|c|}{\textcolor[rgb]{1,0,0}{0.902}}&\multicolumn{2}{|c|}{\textcolor[rgb]{0,1,0}{0.040}}&\multicolumn{2}{|c|}{\textcolor[rgb]{0,0,1}{0.878}}
&\multicolumn{2}{|c|}{\textcolor[rgb]{1,0,0}{0.810}}&\multicolumn{2}{|c|}{\textcolor[rgb]{0,0,1}{0.096}}&\multicolumn{2}{|c|}{0.796}
\\
\multicolumn{4}{|c|}{\textbf{ELD}~\cite{lee2016deep}}
&\multicolumn{2}{|c|}{0.611}&\multicolumn{2}{|c|}{0.092}&\multicolumn{2}{|c|}{0.743}
&\multicolumn{2}{|c|}{0.810}&\multicolumn{2}{|c|}{0.080}&\multicolumn{2}{|c|}{0.839}
&\multicolumn{2}{|c|}{0.776}&\multicolumn{2}{|c|}{0.072}&\multicolumn{2}{|c|}{0.823}
&\multicolumn{2}{|c|}{0.718}&\multicolumn{2}{|c|}{0.123}&\multicolumn{2}{|c|}{0.757}
\\
\multicolumn{4}{|c|}{\textbf{LEGS}~\cite{wang2015deep}}
&\multicolumn{2}{|c|}{0.592}&\multicolumn{2}{|c|}{0.133}&\multicolumn{2}{|c|}{0.701}
&\multicolumn{2}{|c|}{0.785}&\multicolumn{2}{|c|}{0.118}&\multicolumn{2}{|c|}{0.787}
&\multicolumn{2}{|c|}{0.732}&\multicolumn{2}{|c|}{0.118}&\multicolumn{2}{|c|}{0.745}
&\multicolumn{2}{|c|}{--}&\multicolumn{2}{|c|}{--}&\multicolumn{2}{|c|}{--}
\\
\multicolumn{4}{|c|}{\textbf{MDF}~\cite{li2015visual}}
&\multicolumn{2}{|c|}{0.644}&\multicolumn{2}{|c|}{\textcolor[rgb]{0,0,1}{0.092}}&\multicolumn{2}{|c|}{0.703}
&\multicolumn{2}{|c|}{0.807}&\multicolumn{2}{|c|}{0.105}&\multicolumn{2}{|c|}{0.776}
&\multicolumn{2}{|c|}{0.802}&\multicolumn{2}{|c|}{0.095}&\multicolumn{2}{|c|}{0.779}
&\multicolumn{2}{|c|}{0.709}&\multicolumn{2}{|c|}{0.146}&\multicolumn{2}{|c|}{0.692}
\\
\multicolumn{4}{|c|}{\textbf{RFCN}~\cite{wang2016saliency}}
&\multicolumn{2}{|c|}{0.627}&\multicolumn{2}{|c|}{0.111}&\multicolumn{2}{|c|}{0.752}
&\multicolumn{2}{|c|}{0.834}&\multicolumn{2}{|c|}{0.107}&\multicolumn{2}{|c|}{0.852}
&\multicolumn{2}{|c|}{0.838}&\multicolumn{2}{|c|}{0.088}&\multicolumn{2}{|c|}{0.860}
&\multicolumn{2}{|c|}{0.751}&\multicolumn{2}{|c|}{0.132}&\multicolumn{2}{|c|}{0.799}
\\
\multicolumn{4}{|c|}{\textbf{UCF}~\cite{zhang2017learning}}
&\multicolumn{2}{|c|}{0.621}&\multicolumn{2}{|c|}{0.120}&\multicolumn{2}{|c|}{0.748}
&\multicolumn{2}{|c|}{0.844}&\multicolumn{2}{|c|}{0.069}&\multicolumn{2}{|c|}{\textcolor[rgb]{0,0,1}{0.884}}
&\multicolumn{2}{|c|}{0.823}&\multicolumn{2}{|c|}{0.061}&\multicolumn{2}{|c|}{0.874}
&\multicolumn{2}{|c|}{0.735}&\multicolumn{2}{|c|}{0.115}&\multicolumn{2}{|c|}{0.806}
\\
\hline
\hline
\multicolumn{4}{|c|}{\textbf{BL}~\cite{tong2015bootstrap}}
&\multicolumn{2}{|c|}{0.499}&\multicolumn{2}{|c|}{0.239}&\multicolumn{2}{|c|}{0.625}
&\multicolumn{2}{|c|}{0.684}&\multicolumn{2}{|c|}{0.216}&\multicolumn{2}{|c|}{0.714}
&\multicolumn{2}{|c|}{0.666}&\multicolumn{2}{|c|}{0.207}&\multicolumn{2}{|c|}{0.702}
&\multicolumn{2}{|c|}{0.574}&\multicolumn{2}{|c|}{0.249}&\multicolumn{2}{|c|}{0.647}
\\
\multicolumn{4}{|c|}{\textbf{BSCA}~\cite{qin2015saliency}}
&\multicolumn{2}{|c|}{0.509}&\multicolumn{2}{|c|}{0.190}&\multicolumn{2}{|c|}{0.652}
&\multicolumn{2}{|c|}{0.705}&\multicolumn{2}{|c|}{0.182}&\multicolumn{2}{|c|}{0.725}
&\multicolumn{2}{|c|}{0.658}&\multicolumn{2}{|c|}{0.175}&\multicolumn{2}{|c|}{0.705}
&\multicolumn{2}{|c|}{0.601}&\multicolumn{2}{|c|}{0.223}&\multicolumn{2}{|c|}{0.652}
\\
\multicolumn{4}{|c|}{\textbf{DRFI}~\cite{jiang2013salient}}
&\multicolumn{2}{|c|}{0.550}&\multicolumn{2}{|c|}{0.138}&\multicolumn{2}{|c|}{0.688}
&\multicolumn{2}{|c|}{0.733}&\multicolumn{2}{|c|}{0.164}&\multicolumn{2}{|c|}{0.752}
&\multicolumn{2}{|c|}{0.726}&\multicolumn{2}{|c|}{0.145}&\multicolumn{2}{|c|}{0.743}
&\multicolumn{2}{|c|}{0.618}&\multicolumn{2}{|c|}{0.207}&\multicolumn{2}{|c|}{0.670}
\\
\multicolumn{4}{|c|}{\textbf{DSR}~\cite{li2013saliency}}
&\multicolumn{2}{|c|}{0.524}&\multicolumn{2}{|c|}{0.139}&\multicolumn{2}{|c|}{0.660}
&\multicolumn{2}{|c|}{0.662}&\multicolumn{2}{|c|}{0.178}&\multicolumn{2}{|c|}{0.731}
&\multicolumn{2}{|c|}{0.682}&\multicolumn{2}{|c|}{0.142}&\multicolumn{2}{|c|}{0.701}
&\multicolumn{2}{|c|}{0.558}&\multicolumn{2}{|c|}{0.215}&\multicolumn{2}{|c|}{0.594}
\\
\hline
\end{tabular}
\vspace{1mm}
\caption{Quantitative comparisons with 15 methods on 4 large-scale datasets. The best three results are shown in \textcolor[rgb]{1,0,0}{red},~\textcolor[rgb]{0,1,0}{green} and \textcolor[rgb]{0,0,1}{blue}, respectively.
Our method (VGG-16) ranks first or second on these datasets. ``--'' means corresponding methods are trained on that dataset.}
\vspace{-9mm}
\label{table:fauc}
\end{center}
\end{table*}

\textbf{Quantitative Evaluation.}
As illustrated in Fig.~\ref{fig:PR-curve} and Tab.~\ref{table:fauc}, our algorithm with the VGG-16 model already outperforms other competing algorithms across all the datasets in terms of near all evaluation metrics.
Due to the limitation of space, we present the quantitative results on the DUTS-TE, SED and SOD datasets in the supplemental material.
From the results, we have several notable observations: (1) deep learning based methods consistently outperform traditional methods with a large margin, which further proves the superiority of deep features for saliency detection.
(2) DSS~\cite{DSSalCVPR2017}, DCL~\cite{Li2016Deep} and RFCN~\cite{wang2016saliency} are all built on pre-trained segmentation models, \ie, enhanced DeepLab~\cite{chen2017rethinking} and FCNs~\cite{long2015fully},
while our method fine-tuning from image classification models achieves the best results (also better than Amulet~\cite{zhang2017amulet}), especially on the ECSSD and HKU-IS datasets, where our method achieves about 3\% performance leap of F-measure
and around 6\% improvement of S-measure, as well as around 3\% decrease in MAE compared with existing best methods.
(3) Compared to the DHS~\cite{liu2016dhsnet}, DSS~\cite{DSSalCVPR2017} and Amulet~\cite{zhang2017amulet} methods, our method is inferior on several datasets. However, these methods need larger storage space and more computational time.

\begin{figure*}
\centering
\begin{tabular}{@{}c@{}c@{}c@{}c@{}c@{}c@{}c@{}c@{}c@{}c@{}c@{}c}
\vspace{-1mm}
\hspace{-4mm}
\includegraphics[width=0.085\linewidth,height=1.25cm]{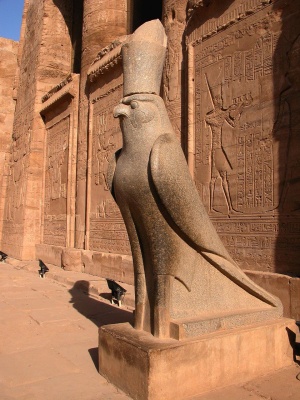}\hspace{0.1mm}\ &
\includegraphics[width=0.085\linewidth,height=1.25cm]{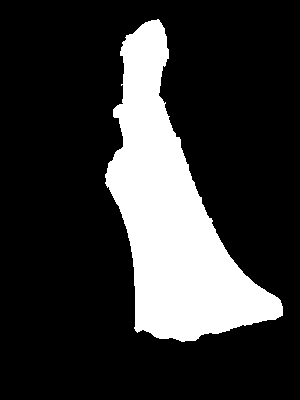}\hspace{0.1mm}\ &
\includegraphics[width=0.085\linewidth,height=1.25cm]{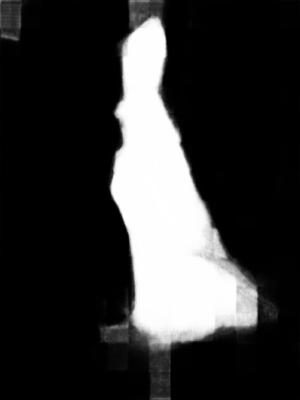}\hspace{0.1mm}\ &
\includegraphics[width=0.085\linewidth,height=1.25cm]{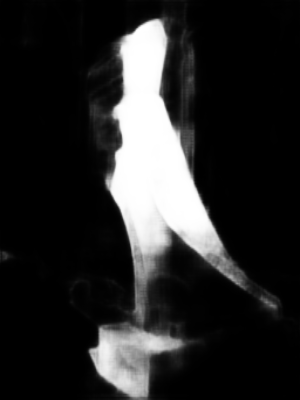}\hspace{0.1mm}\ &
\includegraphics[width=0.085\linewidth,height=1.25cm]{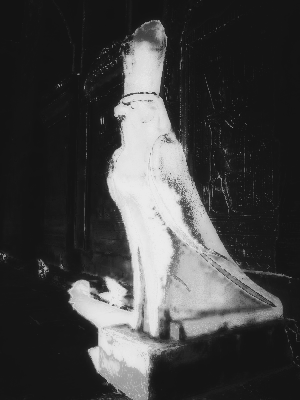}\hspace{0.1mm}\ &
\includegraphics[width=0.085\linewidth,height=1.25cm]{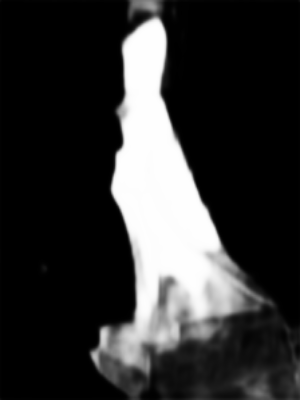}\hspace{0.1mm}\ &
\includegraphics[width=0.085\linewidth,height=1.25cm]{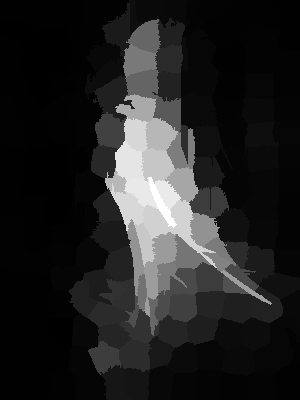}\hspace{0.1mm}\ &
\includegraphics[width=0.085\linewidth,height=1.25cm]{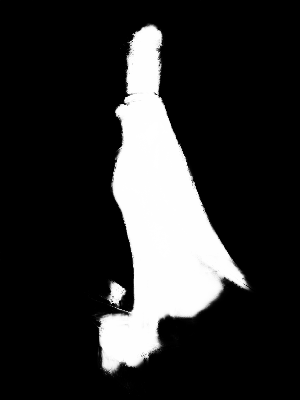}\hspace{0.1mm}\ &
\includegraphics[width=0.085\linewidth,height=1.25cm]{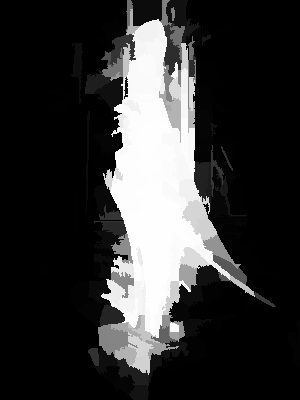}\hspace{0.1mm}\ &
\includegraphics[width=0.085\linewidth,height=1.25cm]{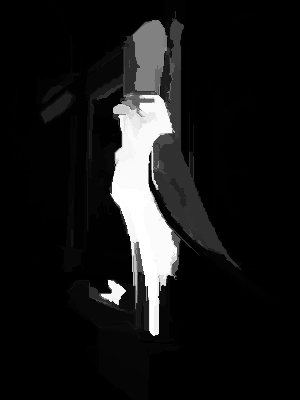}\hspace{0.1mm}\ &
\includegraphics[width=0.085\linewidth,height=1.25cm]{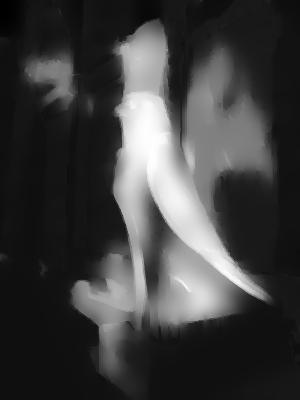}\hspace{0.1mm}\ &
\includegraphics[width=0.085\linewidth,height=1.25cm]{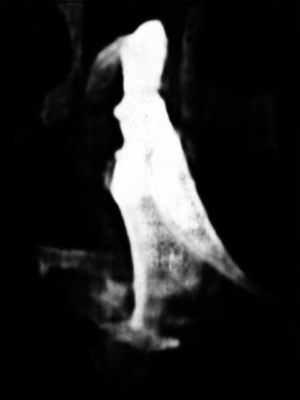}\hspace{0.1mm}\ \\
\vspace{-1mm}
\hspace{-4mm}
\includegraphics[width=0.085\linewidth,height=1.25cm]{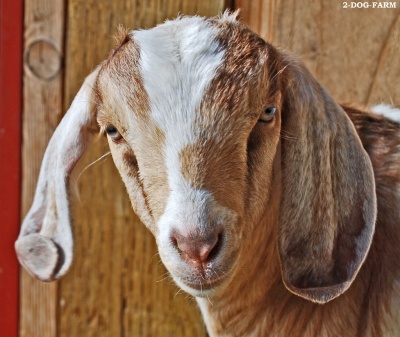}\hspace{0.1mm}\ &
\includegraphics[width=0.085\linewidth,height=1.25cm]{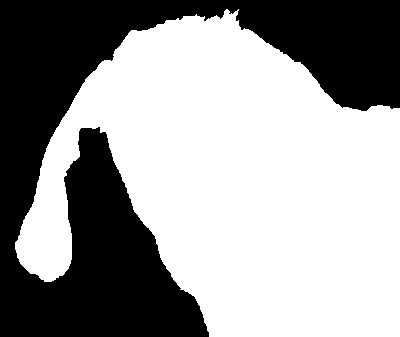}\hspace{0.1mm}\ &
\includegraphics[width=0.085\linewidth,height=1.25cm]{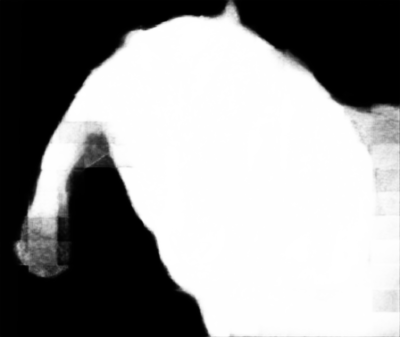}\hspace{0.1mm}\ &
\includegraphics[width=0.085\linewidth,height=1.25cm]{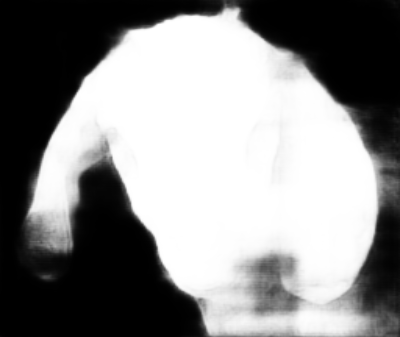}\hspace{0.1mm}\ &
\includegraphics[width=0.085\linewidth,height=1.25cm]{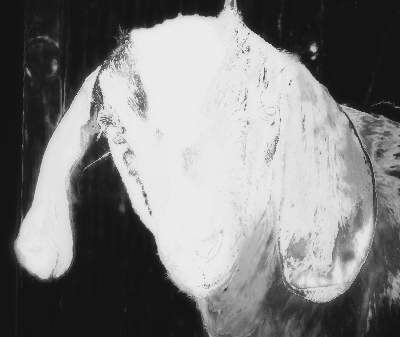}\hspace{0.1mm}\ &
\includegraphics[width=0.085\linewidth,height=1.25cm]{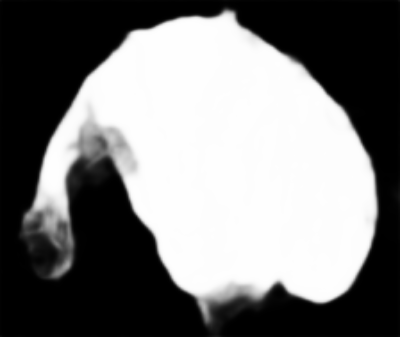}\hspace{0.1mm}\ &
\includegraphics[width=0.085\linewidth,height=1.25cm]{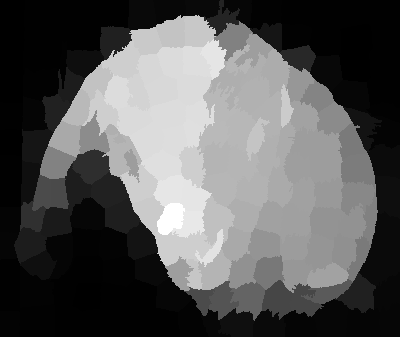}\hspace{0.1mm}\ &
\includegraphics[width=0.085\linewidth,height=1.25cm]{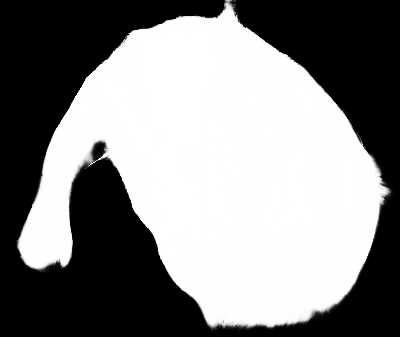}\hspace{0.1mm}\ &
\includegraphics[width=0.085\linewidth,height=1.25cm]{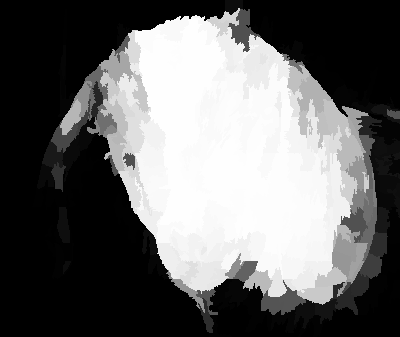}\hspace{0.1mm}\ &
\includegraphics[width=0.085\linewidth,height=1.25cm]{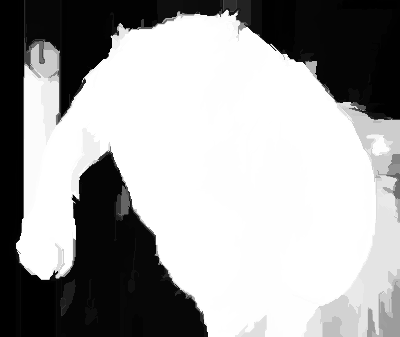}\hspace{0.1mm}\ &
\includegraphics[width=0.085\linewidth,height=1.25cm]{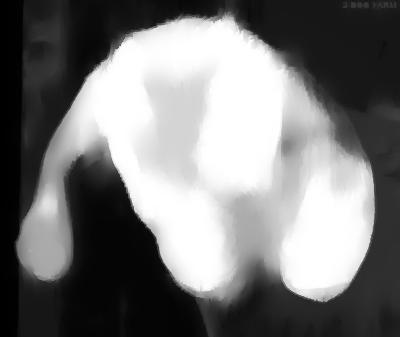}\hspace{0.1mm}\ &
\includegraphics[width=0.085\linewidth,height=1.25cm]{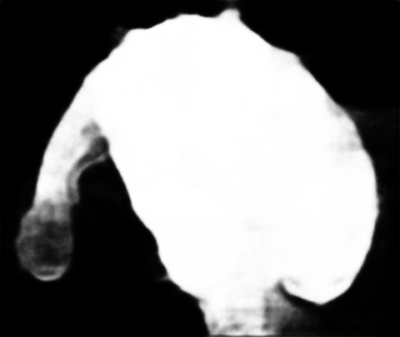}\hspace{0.1mm}\ \\
\vspace{-1mm}
\hspace{-4mm}
\includegraphics[width=0.085\linewidth,height=1.25cm]{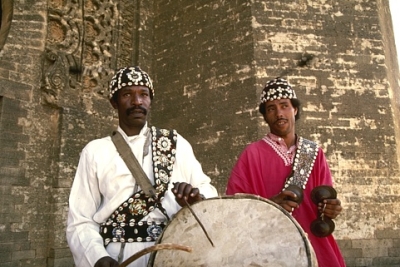}\hspace{0.1mm}\ &
\includegraphics[width=0.085\linewidth,height=1.25cm]{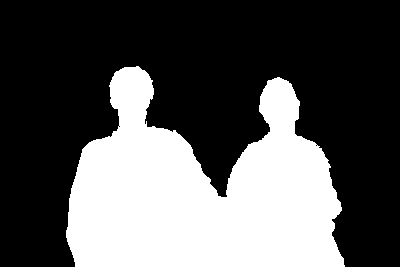}\hspace{0.1mm}\ &
\includegraphics[width=0.085\linewidth,height=1.25cm]{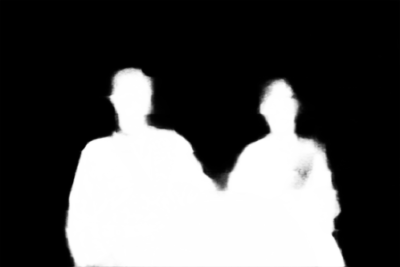}\hspace{0.1mm}\ &
\includegraphics[width=0.085\linewidth,height=1.25cm]{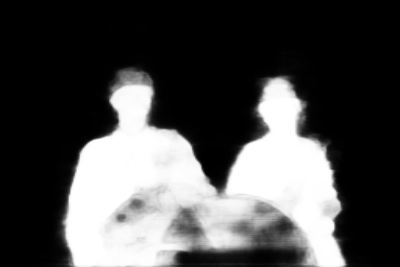}\hspace{0.1mm}\ &
\includegraphics[width=0.085\linewidth,height=1.25cm]{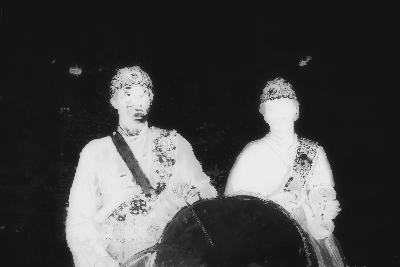}\hspace{0.1mm}\ &
\includegraphics[width=0.085\linewidth,height=1.25cm]{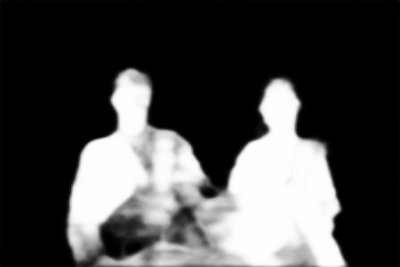}\hspace{0.1mm}\ &
\includegraphics[width=0.085\linewidth,height=1.25cm]{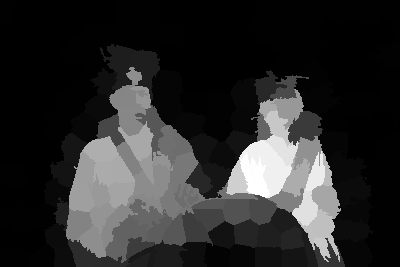}\hspace{0.1mm}\ &
\includegraphics[width=0.085\linewidth,height=1.25cm]{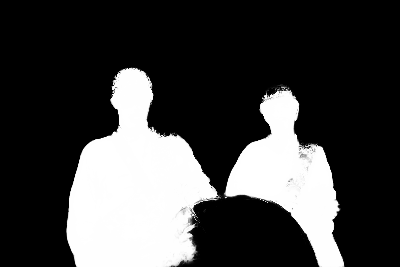}\hspace{0.1mm}\ &
\includegraphics[width=0.085\linewidth,height=1.25cm]{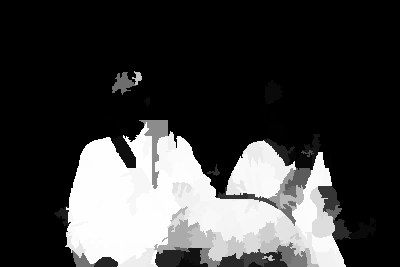}\hspace{0.1mm}\ &
\includegraphics[width=0.085\linewidth,height=1.25cm]{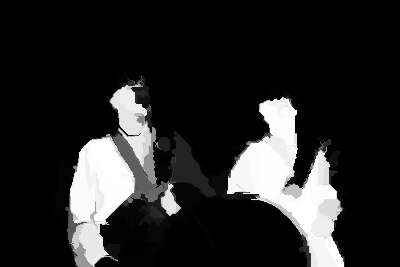}\hspace{0.1mm}\ &
\includegraphics[width=0.085\linewidth,height=1.25cm]{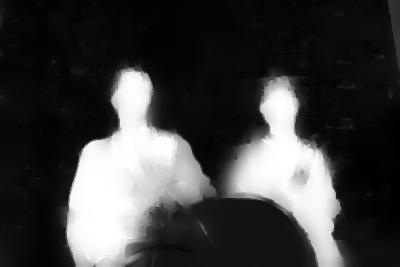}\hspace{0.1mm}\ &
\includegraphics[width=0.085\linewidth,height=1.25cm]{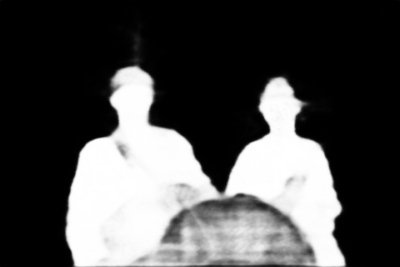}\hspace{0.1mm}\ \\
\vspace{-1mm}
\hspace{-4mm}
\includegraphics[width=0.085\linewidth,height=1.25cm]{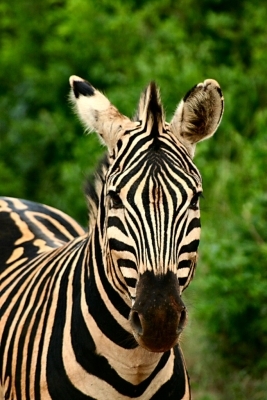}\hspace{0.1mm}\ &
\includegraphics[width=0.085\linewidth,height=1.25cm]{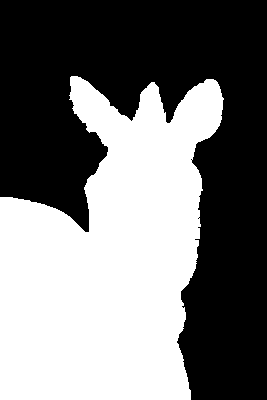}\hspace{0.1mm}\ &
\includegraphics[width=0.085\linewidth,height=1.25cm]{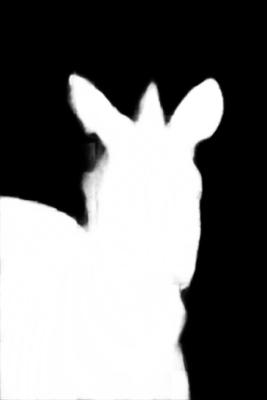}\hspace{0.1mm}\ &
\includegraphics[width=0.085\linewidth,height=1.25cm]{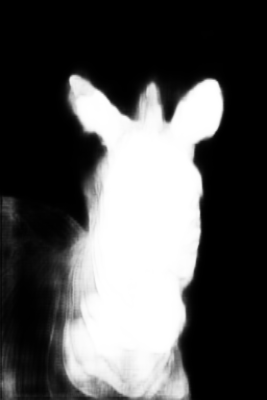}\hspace{0.1mm}\ &
\includegraphics[width=0.085\linewidth,height=1.25cm]{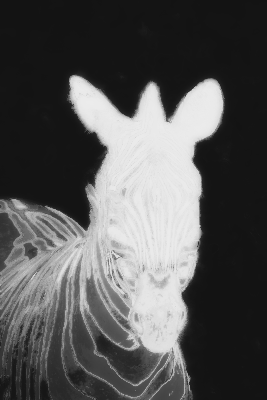}\hspace{0.1mm}\ &
\includegraphics[width=0.085\linewidth,height=1.25cm]{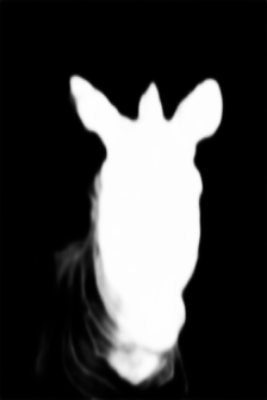}\hspace{0.1mm}\ &
\includegraphics[width=0.085\linewidth,height=1.25cm]{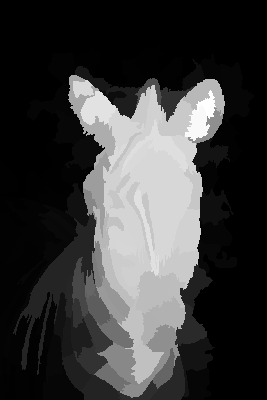}\hspace{0.1mm}\ &
\includegraphics[width=0.085\linewidth,height=1.25cm]{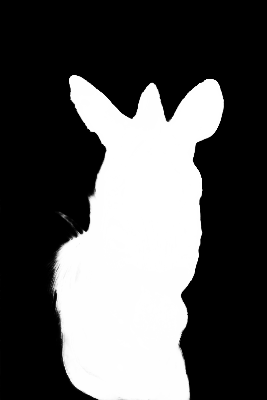}\hspace{0.1mm}\ &
\includegraphics[width=0.085\linewidth,height=1.25cm]{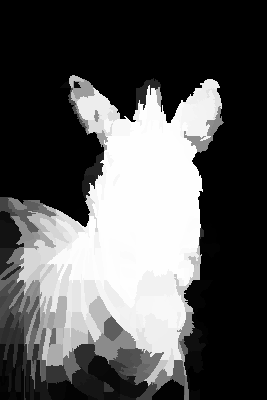}\hspace{0.1mm}\ &
\includegraphics[width=0.085\linewidth,height=1.25cm]{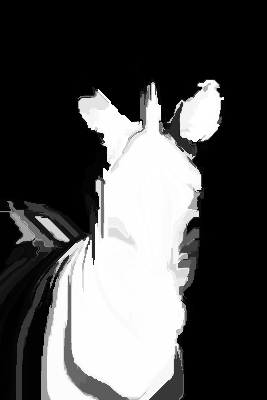}\hspace{0.1mm}\ &
\includegraphics[width=0.085\linewidth,height=1.25cm]{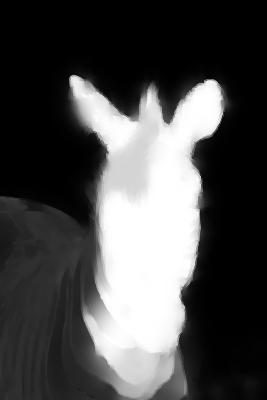}\hspace{0.1mm}\ &
\includegraphics[width=0.085\linewidth,height=1.25cm]{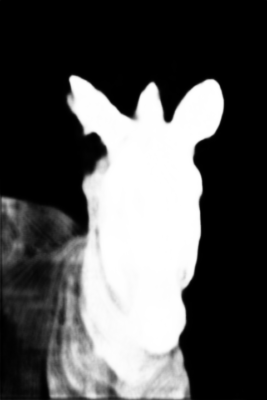}\hspace{0.1mm}\ \\
\vspace{-1mm}
\hspace{-4mm}
\includegraphics[width=0.085\linewidth,height=1.25cm]{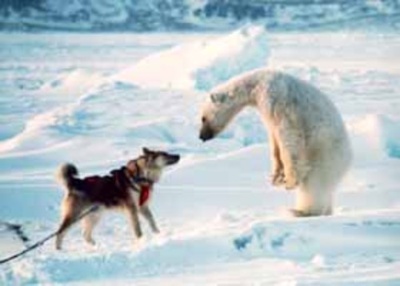}\hspace{0.1mm}\ &
\includegraphics[width=0.085\linewidth,height=1.25cm]{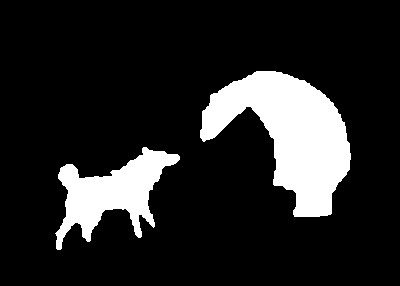}\hspace{0.1mm}\ &
\includegraphics[width=0.085\linewidth,height=1.25cm]{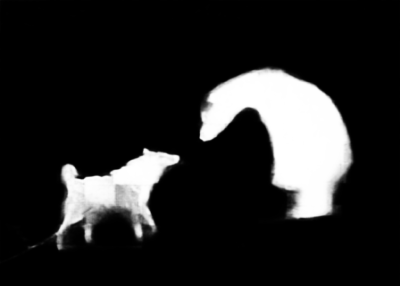}\hspace{0.1mm}\ &
\includegraphics[width=0.085\linewidth,height=1.25cm]{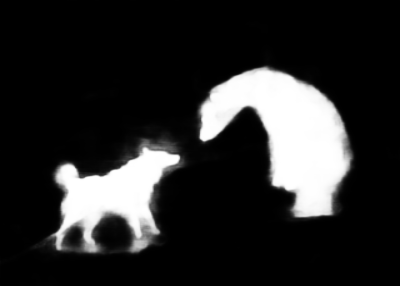}\hspace{0.1mm}\ &
\includegraphics[width=0.085\linewidth,height=1.25cm]{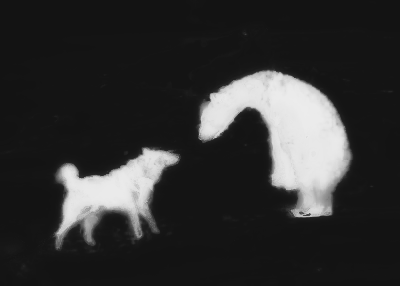}\hspace{0.1mm}\ &
\includegraphics[width=0.085\linewidth,height=1.25cm]{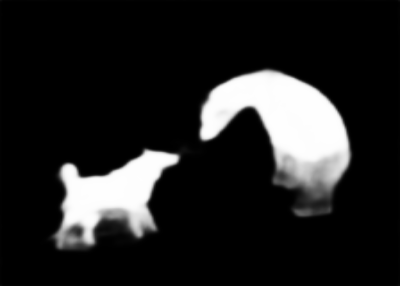}\hspace{0.1mm}\ &
\includegraphics[width=0.085\linewidth,height=1.25cm]{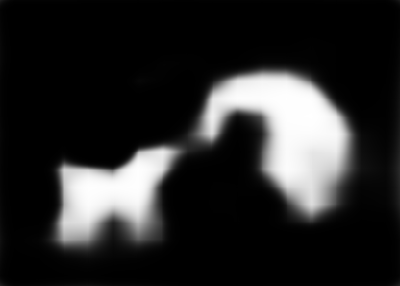}\hspace{0.1mm}\ &
\includegraphics[width=0.085\linewidth,height=1.25cm]{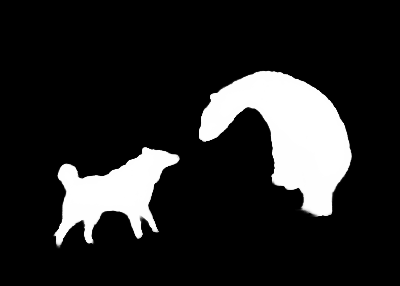}\hspace{0.1mm}\ &
\includegraphics[width=0.085\linewidth,height=1.25cm]{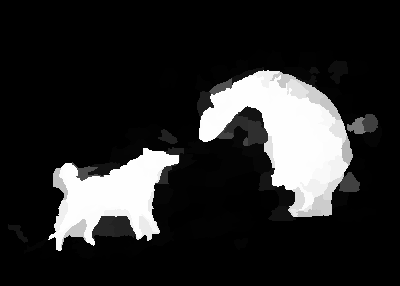}\hspace{0.1mm}\ &
\includegraphics[width=0.085\linewidth,height=1.25cm]{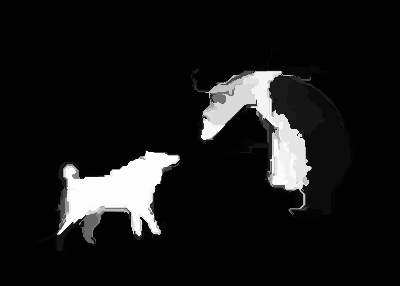}\hspace{0.1mm}\ &
\includegraphics[width=0.085\linewidth,height=1.25cm]{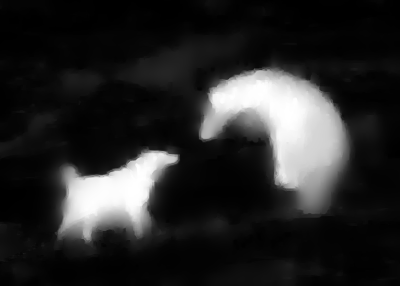}\hspace{0.1mm}\ &
\includegraphics[width=0.085\linewidth,height=1.25cm]{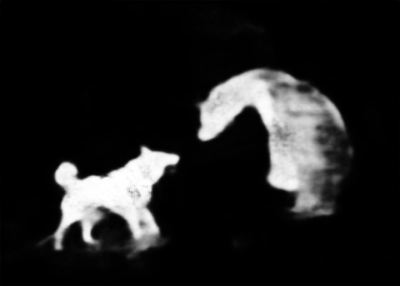}\hspace{0.1mm}\ \\
\vspace{-1mm}
\hspace{-4mm}
\includegraphics[width=0.085\linewidth,height=1.25cm]{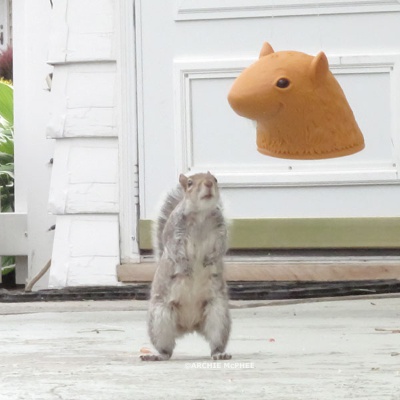}\hspace{0.1mm}\ &
\includegraphics[width=0.085\linewidth,height=1.25cm]{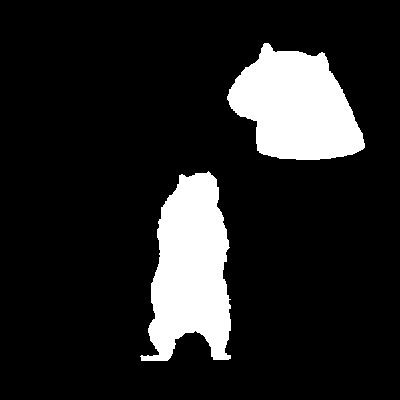}\hspace{0.1mm}\ &
\includegraphics[width=0.085\linewidth,height=1.25cm]{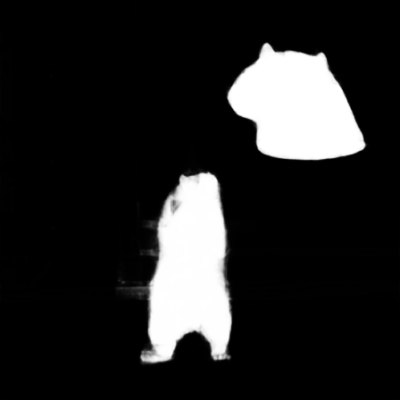}\hspace{0.1mm}\ &
\includegraphics[width=0.085\linewidth,height=1.25cm]{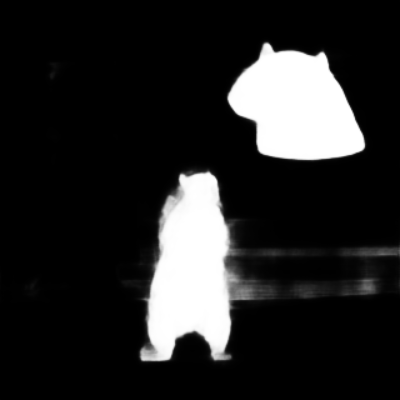}\hspace{0.1mm}\ &
\includegraphics[width=0.085\linewidth,height=1.25cm]{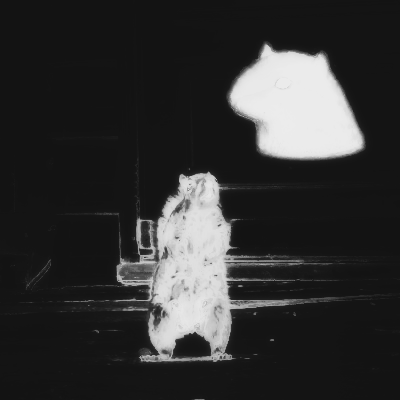}\hspace{0.1mm}\ &
\includegraphics[width=0.085\linewidth,height=1.25cm]{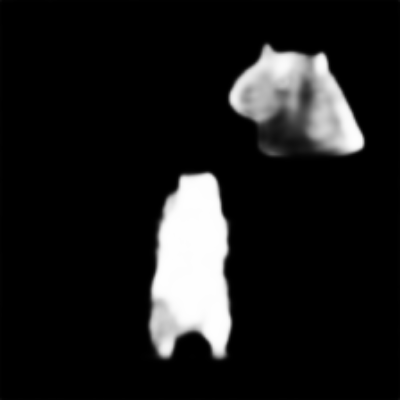}\hspace{0.1mm}\ &
\includegraphics[width=0.085\linewidth,height=1.25cm]{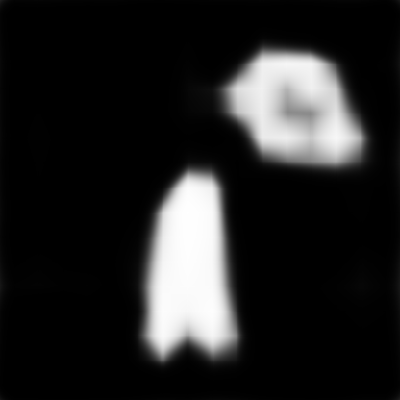}\hspace{0.1mm}\ &
\includegraphics[width=0.085\linewidth,height=1.25cm]{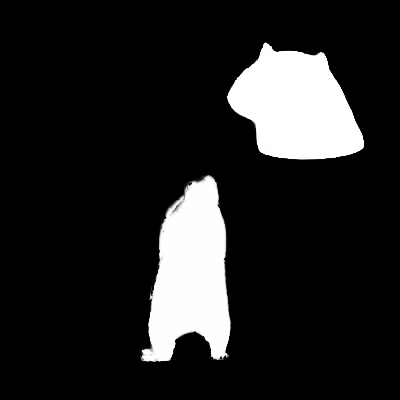}\hspace{0.1mm}\ &
\includegraphics[width=0.085\linewidth,height=1.25cm]{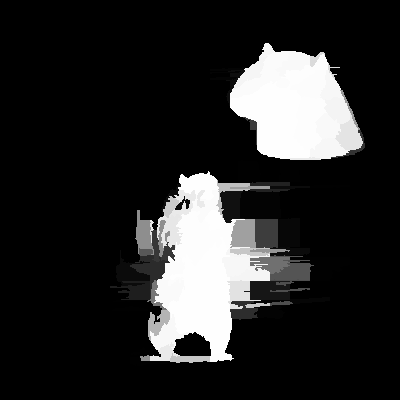}\hspace{0.1mm}\ &
\includegraphics[width=0.085\linewidth,height=1.25cm]{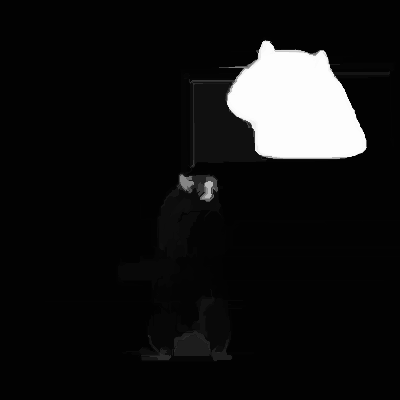}\hspace{0.1mm}\ &
\includegraphics[width=0.085\linewidth,height=1.25cm]{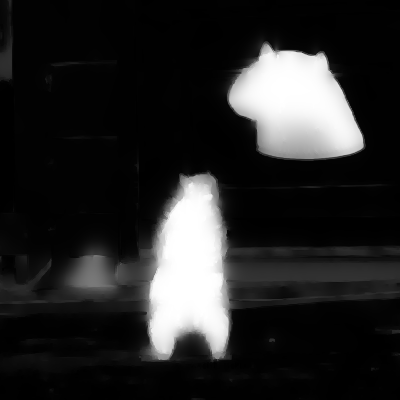}\hspace{0.1mm}\ &
\includegraphics[width=0.085\linewidth,height=1.25cm]{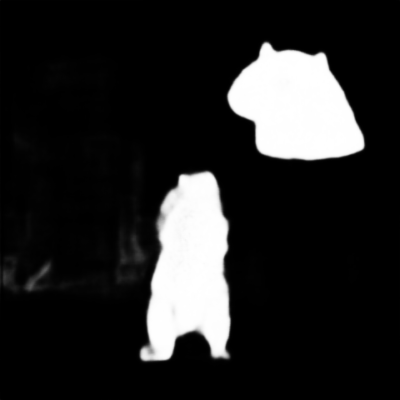}\hspace{0.1mm}\ \\
{\small (a)} & {\small(b)} & {\small(c)} & {\small(d)} & {\small(e)}& {\small(f)}& {\small(g)}
& {\small(h)}& {\small(i)}& {\small(j)}& {\small(k)}& {\small(l)}\ \\
\end{tabular}
\caption{Comparison of saliency maps. (a) Input images; (b) Ground truth; (c) Ours; (d) Amulet~\cite{zhang2017amulet}; (e) DCL~\cite{Li2016Deep}; (f) DHS~\cite{liu2016dhsnet}; (g) DS~\cite{Li2016DeepSaliency}; (h) DSS~\cite{DSSalCVPR2017}; (i) ELD~\cite{lee2016deep}; (j) MDF~\cite{li2015visual}; (k) RFCN~\cite{wang2016saliency}; (l) UCF~\cite{zhang2017learning}. Due to the limitation of space, we don't show the results of LEGS~\cite{wang2015deep}, BL~\cite{tong2015bootstrap}, BSCA~\cite{qin2015saliency}, DRFI~\cite{jiang2013salient} and DSR~\cite{li2013saliency}. The results can be found in the supplemental material.
\label{fig:map_comparison}}
\vspace{-2mm}
\end{figure*}

\textbf{Qualitative Evaluation.}
Fig.~\ref{fig:map_comparison} provides several visual comparisons, where our method outperforms
the compared methods in various challenging cases.
For example, the images in the first two rows are very low contrast, where most of the compared methods fail
to capture the salient objects, while our method successfully highlights them with sharper edge preserved.
Salient objects in the 2-4 rows are near the image boundary, and most of the compared methods can not predict the whole objects, while our method captures the whole salient regions with high precision.
Images in the 5-6 rows have multiple disconnected salient objects, which lead to false detection especially for ELD~\cite{lee2016deep}, MDF~\cite{li2015visual} and RFCN~\cite{wang2016saliency}, and our method achieves consistently better results in this challenging case.

\textbf{Ablation Studies.}
To analyze the role of different components in our model, we perform the following ablation studies on the ECSSD, HKU-IS and PASCAL-S datasets.
1) To validate the effectiveness of our aggregation method, we run two baselines with the VGG-16 model.
For the first one (Tab.~\ref{table:attention} model (a)), we directly use the side-out features and add the binary classifiers to the model, similar to the HED~\cite{xie2015holistically}.
We train this model to analyze whether the aggregated features (Tab.~\ref{table:attention} model (b)) can lead to better performance.
2) We also use the bottom-up attention (from low-layer to high-layer) to train our framework (Tab.~\ref{table:attention} model (c)).
This model is used to verify whether our model trained with attention in the opposite direction can help to predict good results.
3) To verify the effects of contextual pyramids, we additionally train a model with single attention (Tab.~\ref{table:attention} model (d)).
The resulting model (Tab.~\ref{table:attention} model (e)) is used for our results in Tab.~\ref{table:fauc}.
4) In addition, we build our framework with the ResNet-50 model (Tab.~\ref{table:attention} model (f)).
This model is used to prove that our method can consistently boost the saliency accuracy with more powerful features.
Results are aslo shown in Tab.~\ref{table:attention}.
Comparing the results of the model (a) and model (b), we find that the aggregated features greatly improve the performance, which convincingly demonstrates the effectiveness of the proposed method.
The model (c) shows no advantage over model (b), indicating that bottom-up attention is not effective in our framework because deep CNNs already have intrinsic properties of the bottom-up feature extraction.
However, using top-down attention with the aggregated features (model (d)) improves the performance with a large margin (4\% increase over the baseline (model (a)).
In addition, the performance is further boosted by using contextual attention pyramids.
%
\setlength{\tabcolsep}{5pt}
\begin{table*}
\doublerulesep=0.6pt
\begin{center}
\begin{tabular}{|c|c|c|c|c|c|c|c|c|c|c|c|c|c|c|c|c|c|c|c|c|c|c|c|c|||c|c|c|c|c|c|c|c|||}
\hline
\multicolumn{4}{|c|}{}
&\multicolumn{6}{|c|}{ECSSD~\cite{yan2013hierarchical}}
&\multicolumn{6}{|c|}{HKU-IS~\cite{li2015visual}}
&\multicolumn{6}{|c|}{PASCAL-S~\cite{li2014secrets}}
\\
\hline
\hline
\multicolumn{4}{|c|}{Models}
&\multicolumn{2}{|c|}{$F_\eta$}&\multicolumn{2}{|c|}{$MAE$}&\multicolumn{2}{|c|}{$S_\lambda$}
&\multicolumn{2}{|c|}{$F_\eta$}&\multicolumn{2}{|c|}{$MAE$}&\multicolumn{2}{|c|}{$S_\lambda$}
&\multicolumn{2}{|c|}{$F_\eta$}&\multicolumn{2}{|c|}{$MAE$}&\multicolumn{2}{|c|}{$S_\lambda$}
\\
\hline
\hline
\multicolumn{4}{|c|}{\footnotesize{(a):side-out features (VGG-16)}}
&\multicolumn{2}{|c|}{0.805}&\multicolumn{2}{|c|}{0.131}&\multicolumn{2}{|c|}{0.813}
&\multicolumn{2}{|c|}{0.781}&\multicolumn{2}{|c|}{0.108}&\multicolumn{2}{|c|}{0.820}
&\multicolumn{2}{|c|}{0.712}&\multicolumn{2}{|c|}{0.157}&\multicolumn{2}{|c|}{0.756}
\\
\hline
\multicolumn{4}{|c|}{\footnotesize{(b):aggregated features (VGG-16)}}
&\multicolumn{2}{|c|}{0.824}&\multicolumn{2}{|c|}{0.120}&\multicolumn{2}{|c|}{0.821}
&\multicolumn{2}{|c|}{0.803}&\multicolumn{2}{|c|}{0.085}&\multicolumn{2}{|c|}{0.841}
&\multicolumn{2}{|c|}{0.731}&\multicolumn{2}{|c|}{0.140}&\multicolumn{2}{|c|}{0.773}
\\
\hline
\multicolumn{4}{|c|}{\footnotesize{(c):(b)+bottom-up single attention (VGG-16)}}
&\multicolumn{2}{|c|}{0.837}&\multicolumn{2}{|c|}{0.112}&\multicolumn{2}{|c|}{0.817}
&\multicolumn{2}{|c|}{0.805}&\multicolumn{2}{|c|}{0.080}&\multicolumn{2}{|c|}{0.846}
&\multicolumn{2}{|c|}{0.740}&\multicolumn{2}{|c|}{0.135}&\multicolumn{2}{|c|}{0.780}
\\
\hline
\multicolumn{4}{|c|}{\footnotesize{(d):(b)+ top-down single attention (VGG-16)}}
&\multicolumn{2}{|c|}{0.845}&\multicolumn{2}{|c|}{0.093}&\multicolumn{2}{|c|}{0.876}
&\multicolumn{2}{|c|}{0.827}&\multicolumn{2}{|c|}{0.078}&\multicolumn{2}{|c|}{0.852}
&\multicolumn{2}{|c|}{0.752}&\multicolumn{2}{|c|}{0.128}&\multicolumn{2}{|c|}{0.800}
\\
\hline
\multicolumn{4}{|c|}{\footnotesize{(e):(b)+top-down attention pyramid (VGG-16)}}
&\multicolumn{2}{|c|}{0.887}&\multicolumn{2}{|c|}{0.049}&\multicolumn{2}{|c|}{0.902}
&\multicolumn{2}{|c|}{0.861}&\multicolumn{2}{|c|}{\textbf{0.038}}&\multicolumn{2}{|c|}{0.891}
&\multicolumn{2}{|c|}{0.794}&\multicolumn{2}{|c|}{0.092}&\multicolumn{2}{|c|}{0.832}
\\
\hline
\hline
\multicolumn{4}{|c|}{\footnotesize{(f):(b)+top-down attention pyramid (ResNet-50)}}
&\multicolumn{2}{|c|}{\textbf{0.912}}&\multicolumn{2}{|c|}{\textbf{0.042}}&\multicolumn{2}{|c|}{\textbf{0.923}}
&\multicolumn{2}{|c|}{\textbf{0.887}}&\multicolumn{2}{|c|}{0.039}&\multicolumn{2}{|c|}{\textbf{0.907}}
&\multicolumn{2}{|c|}{\textbf{0.823}}&\multicolumn{2}{|c|}{\textbf{0.084}}&\multicolumn{2}{|c|}{\textbf{0.844}}
\\
\hline
\end{tabular}
\end{center}
\vspace{-2mm}
\caption{Experimental results using different model settings, evaluated on the ECSSD, HKU-IS and PASCAL-S datasets. All models are trained on the augmented MSRA10K dataset and share the same hyper-parameters described in subsection 5.1.}
\vspace{-5mm}
\label{table:attention}
\end{table*}
\vspace{-5mm}
\subsubsection{Runtime Testing and Analysis}
\setlength{\tabcolsep}{5pt}
\begin{table}
\begin{center}
\begin{tabular}{|c|c|c|c|c|c|c|}
\hline
Settings&DHS&DSS&Amulet&Ours\\
\hline
Train time (h)     &--&--&107&\textbf{22} \\
\hline
Test rate (fps) GPU&21.4&2.1&15.4&\textbf{30.2}\\
\hline
Test rate (fps) CPU&0.095&0.054&0.082&\textbf{3.51} \\
\hline
Model size (MB)    &358&237&126&\textbf{67}\\
\hline
\end{tabular}
\end{center}
\vspace{-2mm}
\caption{Comparison of the runtime and model size.}
\label{table:time}
\vspace{-5mm}
\end{table}
~~Another advantage of our method is the fast training and real-time testing.
Tab.~\ref{table:time} shows a comparison of training time (hours), testing rate ( frames per second), and binarized model size (MB) between DHS, DSS, Amulet and our method.
All models were tested with $256\times256\times3$ images.
Times were measured in a quad-core PC machine with an i5-6600 CPU and an NVIDIA Titan 1080 GPU (with 8G memory).
For the same VGG16-model, our method processes images twice faster than Amulet.
Training time is reduced by 5, from 107 hours to 22.
The main reason is the introduction of the contextual attention module, which rapidly highlight the salient objects or regions on the convolutional feature maps during the network training.
In other words, the proposed  contextual attention expedites object-related feature learning.
In addition, our aggregating feature method significantly reduces the model size.
The small model size (67MB) also makes our model faster in the testing.
Our method run 30 \emph{fps} and is faster than other methods.
The real-time speed will foster more applications.
\vspace{-5mm}
\section{Conclusion}

In this paper, we propose an agile aggregating multi-level feature framework for salient object detection.
We introduces a contextual attention module between convolutional layers.
It is effective for localizing the salient objects and guiding low-layer feature learning.
We also propose a new aggregating feature method only using the side-outputs of pre-trained backbone networks.
The new method is realized with minimal complexity, requiring fewer parameters than the previous one in Amulet.
Extensive experiments demonstrate that our method performs favorably against state-of-the-art saliency approaches in both accuracy and speed.

{\footnotesize
\bibliographystyle{ieee}
\bibliography{egbib}
}

\end{document}